\documentclass{article}

\usepackage[preprint]{neurips_data_2023}

\usepackage[utf8]{inputenc} 
\usepackage[T1]{fontenc}    
\usepackage{hyperref}       
\usepackage{url}            
\usepackage{booktabs}       
\usepackage{amsfonts}       
\usepackage{nicefrac}       
\usepackage{microtype}      
\usepackage{xcolor}         
\usepackage{graphicx}

\newcommand{\name}{HumanEdit}
\usepackage{multirow} 

\usepackage{xcolor}
\usepackage{pifont}
\definecolor{my_green}{RGB}{51,102,0}
\definecolor{my_red}{RGB}{204, 0, 0}
\renewcommand{\checkmark}{\textcolor{my_green}{\ding{51}}} 
\newcommand{\crossmark}{\textcolor{my_red}{\ding{55}}} 
\usepackage{float}
\usepackage{marvosym}

\title{\name: A High-Quality Human-Rewarded Dataset for Instruction-based Image Editing} 

\author{%
Jinbin Bai$^{1,2*}$\quad Wei Chow$^{1*}$\quad Ling Yang$^{3}$\quad Xiangtai Li$^{2}$\quad \\
\textbf{Juncheng Li}$^{1}$\quad \textbf{Hanwang Zhang}$^{2,4}$\quad \textbf{Shuicheng Yan}$^{1,2\dag}$\vspace{0.5em}\\
$^1$National University of Singapore\quad $^2$Skywork AI\quad \\$^3$Peking University\quad $^4$Nanyang Technological University \vspace{0.5em}\\
\texttt{* Equal contributions, \quad \dag~Corresponding author} \vspace{0.15cm}\\
  {\centering \texttt{Project Page: \url{https://viiika.github.io/HumanEdit}}}
}

\begin{document}

\maketitle

\begin{figure}[htbp]
	\centering  
	\vspace{-1.5mm}
	\includegraphics[height=6.1cm]{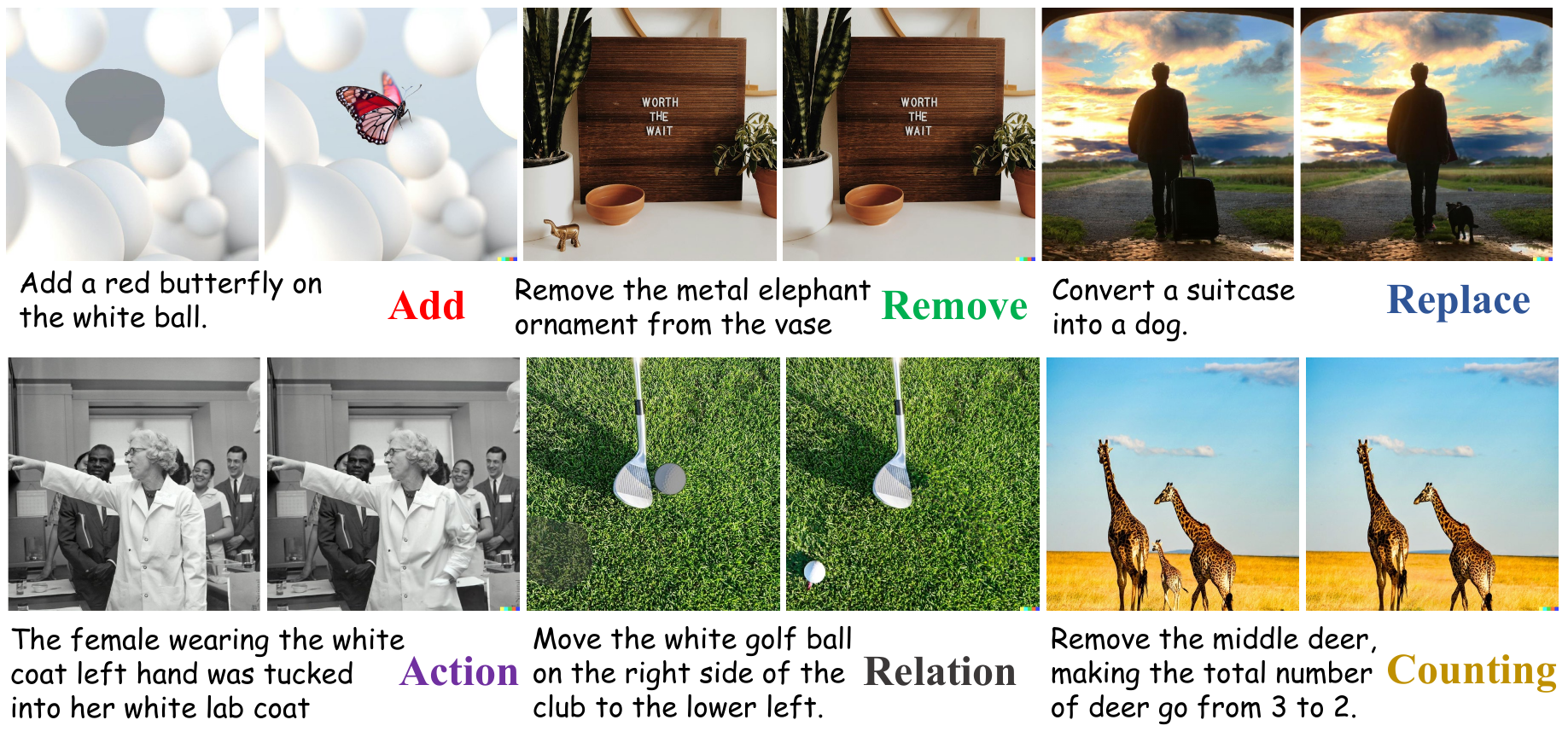}
	\vspace{-2mm}
    	\caption{\textbf{Data examples of instruction-guided image editing in \name}. Our dataset encompasses six distinct editing categories. In the images, gray shapes represent masks, which are provided for every photograph. Moreover, approximately half of the dataset includes instructions that are sufficiently detailed to enable editing without masks. It is important to note that, for conciseness, masks are depicted directly on the original images within this paper; however, in the dataset, the original images and masks are stored separately.}
	\label{fig:case01}  
\end{figure}

\begin{abstract}
We present \name, a high-quality, human-rewarded dataset specifically designed for instruction-guided image editing, enabling precise and diverse image manipulations through open-form language instructions. Previous large-scale editing datasets often incorporate minimal human feedback, leading to challenges in aligning datasets with human preferences. \name{} bridges this gap by employing human annotators to construct data pairs and administrators to provide feedback. With meticulously curation, \name{} comprises 5,751 images and requires more than 2,500 hours of human effort across four stages, ensuring both accuracy and reliability for a wide range of image editing tasks. The dataset includes six distinct types of editing instructions: Action, Add, Counting, Relation, Remove, and Replace, encompassing a broad spectrum of real-world scenarios. All images in the dataset are accompanied by masks, and for a subset of the data, we ensure that the instructions are sufficiently detailed to support mask-free editing. Furthermore, \name{} offers comprehensive diversity and high-resolution $1024 \times 1024$ content sourced from various domains, setting a new versatile benchmark for instructional image editing datasets. With the aim of advancing future research and establishing evaluation benchmarks in the field of image editing, we release \name{} at \url{https://huggingface.co/datasets/BryanW/\name}.
\end{abstract}

\section{Introduction}

In the fields of computer vision and graphics, image-to-image synthesis has been a foundational topic of research for many years. Pioneering works such as CycleGAN~\citep{zhu2017unpaired}, CartoonGAN~\citep{chen2018cartoongan,wang2020learning}, and StyleGAN~\citep{karras2019stylebasedgeneratorarchitecturegenerative} have achieved remarkable success in tasks ranging from unpaired image-to-image translation to high-quality image synthesis. Recent advancements in diffusion models~\citep{sd1.5,podell2023sdxl,sd3} have propelled text-to-image generation to unprecedented levels, largely due to the availability of massive datasets like LAION-5B~\citep{schuhmann2022laion5bopenlargescaledataset}, which provide the necessary scale and diversity for training state-of-the-art models. Building upon these exceptional text-to-image foundation models, numerous works have extended their applications to image-to-image editing~\citep{brooks2023instructpix2pix,bai2023integrating,feng2024item}, video generation~\citep{blattmann2023align,tian2024videotetris,yang2024crossmodal} , 3D generation~\citep{yi2024diffusion,wu2024consistent3d,yi2024mvgamba,yi2023invariant}, and more.

A critical task within image-to-image synthesis is applying semantic edits to specific regions of images. Such operations, categorized as Local Editing~\citep{yu2024anyeditmasteringunifiedhighquality}, are exemplified by works like InstructPix2Pix~\citep{brooks2023instructpix2pix}, which enables image editing based on textual instructions. The increasing demand for precise image editing has driven the creation of specialized datasets~\citep{yang2024editworld,ge2024seed,zhang2024magicbrush}, enabling fine-grained tasks such as style modifications~\citep{zhang2017style}, object changes~\citep{bai2023integrating,feng2024item,shi2024relationbooth,zhou2024magictailor}, and background alterations~\citep{zhang2024transparent}. 

Recently, a number of instruction-based image editing datasets and models have been introduced to advance the performance of models in local editing tasks, such as EMU-Edit~\citep{sheynin2024emu}, HQ-Edit~\citep{hui2024hq}, SEED-Data-Edit~\citep{ge2024seed}, EditWorld~\citep{yang2024editworld}, UltraEdit~\citep{zhao2024ultraedit}, and AnyEdit~\citep{yu2024anyeditmasteringunifiedhighquality}. Despite their contributions, most of these datasets are constructed with image synthesis models and large language models, incorporating minimal human feedback. Consequently, these datasets often fall short of practical applicability. A key challenge lies in \textbf{aligning datasets with human preference}, as the distribution of training data tends to be noisy and misaligned with real-world user editing instructions. This discrepancy gives rise to several issues in image editing tasks. For instance, the phrasing of editing instructions and the mask regions often fail to reflect actual user needs, and the edited outputs frequently exhibit artifacts or inconsistencies with human performance (\textit{e.g.}, body distortions). These intrinsic dataset biases are difficult to address solely through improvements in model
architectures and training schedule. Although datasets like MagicBrush~\citep{zhang2024magicbrush} attempt to address this gap by employing human annotators, they suffer from limitations in image quality and resolution due to the constraints of their original image sources. These shortcomings hinder their ability to support high-quality and high-resolution editing scenarios effectively. We provide a detailed discussion of these limitations in Section~\ref{sec:stat}.

\begin{table}[htbp]
    \small
    \centering
    \caption{
        Distribution of 6 types of our human-rewarded editing instructions.
    }
    \label{tab:distribution}
    \resizebox{.85\columnwidth}{!}{
        \begin{tabular}{cccccccc}
        \toprule
            & Add & Rmove & Replace & Action & Counting & Relation & Sum\\
        \toprule
        \name-full      & 801  & 1,813  & 1,370  & 659  &  698 & 410 &  5,751  \\
        \name-core      & 30  & 188  &  97 & 37  & 20  & 28 &  400  \\     
        \toprule
        \end{tabular}}

\end{table}

Recognizing the importance of addressing these challenges in training dataset to advance instructional image-to-image translation, we introduce \textbf{\name{}}, a high-quality instructional image editing dataset featuring human-annotated instructions. \name{} includes 5,751 high-quality image pairs, each accompanied by editing instructions and detailed image descriptions, and spans six editing categories: \textit{Action}, \textit{Add}, \textit{Counting}, \textit{Relation}, \textit{Remove}, and \textit{Replace} (Tab.~\ref{tab:distribution}). \name{} offers several advantages:

\begin{itemize}
    \item \textbf{Enhanced Data Quality:} Through multi-round quality control, \name{} achieves higher data accuracy and consistency compared to existing datasets. The dataset underwent multiple rounds of validation and modification, totaling approximately 2,500 hours of effort, ensuring suitability for fine-tuning or evaluation benchmarks.
    \item \textbf{Diverse and High-Resolution Sources:} Unlike MagicBrush~\citep{zhang2024magicbrush}, which is limited to the COCO dataset~\citep{lin2014microsoft}, \name{} is sourced from a broader range of origins and includes higher-resolution images, catering to high-fidelity, photo-realistic editing tasks. 
    \item \textbf{Mask Differentiation:} \name{} categorizes images into those requiring masks and those that do not, providing masks where necessary to support diverse fine-tuning and evaluation needs.
    \item \textbf{Increased Diversity:} Analyses such as word cloud visualizations, Vendi Score calculations, sunburst charts, river charts and categorizations of image pair types underscore the dataset's superior diversity.
    \item \textbf{Categorization Across Dimensions:}By classifying editing tasks into six distinct dimensions, \name{} provides a clear framework for evaluation and development.
\end{itemize}

We detail the four-stage annotation pipeline in Section~\ref{sec:pipe} and present dataset statistics in Section~\ref{sec:stat} and Appendix~\ref{app:figures}, including sunburst charts, river charts, and categorizations of image pair types. A guidance book is provided in Appendix~\ref{app:book} for future research reference, along with failure cases excluded from \name{} in Appendix~\ref{app:failure}.

To provide the performance benchmark on \name{} for future evaluation and development, we report several baselines in both mask-free and mask-provided settings in Section~\ref{benchmark}, including InstructPix2Pix~\citep{brooks2023instructpix2pix}, MGIE~\citep{fu2023guiding}, HIVE~\citep{zhang2024hive}, MagicBrush~\citep{zhang2024magicbrush}, Blended Latent Diffusion~\citep{avrahami2023blended}, GLIDE~\citep{nichol2021glide}, aMUSEd~\citep{patil2024amused} and Meissonic~\citep{bai2024meissonic}. Default hyperparameters are used to ensure reproducibility and fairness. And we draw some conclusions, for example, most methods perform better on \textit{Add} tasks than on \textit{Remove} tasks. Mask-provided methods generally achieve superior performance in semantic-level evaluation metrics compared to pixel-level metrics. Furthermore, even for \textit{Add} tasks, challenges persist in cases requiring domain-specific knowledge or handling unfamiliar instructions, such as \textit{``Add a petal in the middle of the white puppy's forehead.”}

This dataset establishes a \textbf{benchmark} for future research, fostering the development of advanced image-to-image translation and editing models.

\section{Dataset Annotation Pipeline}\label{sec:pipe}

The data collection process is outlined in Figure~\ref{fig:pipeline}, which divides the workflow into four distinct stages:

\begin{figure}[h!]
\centering
\includegraphics[width=1\textwidth]{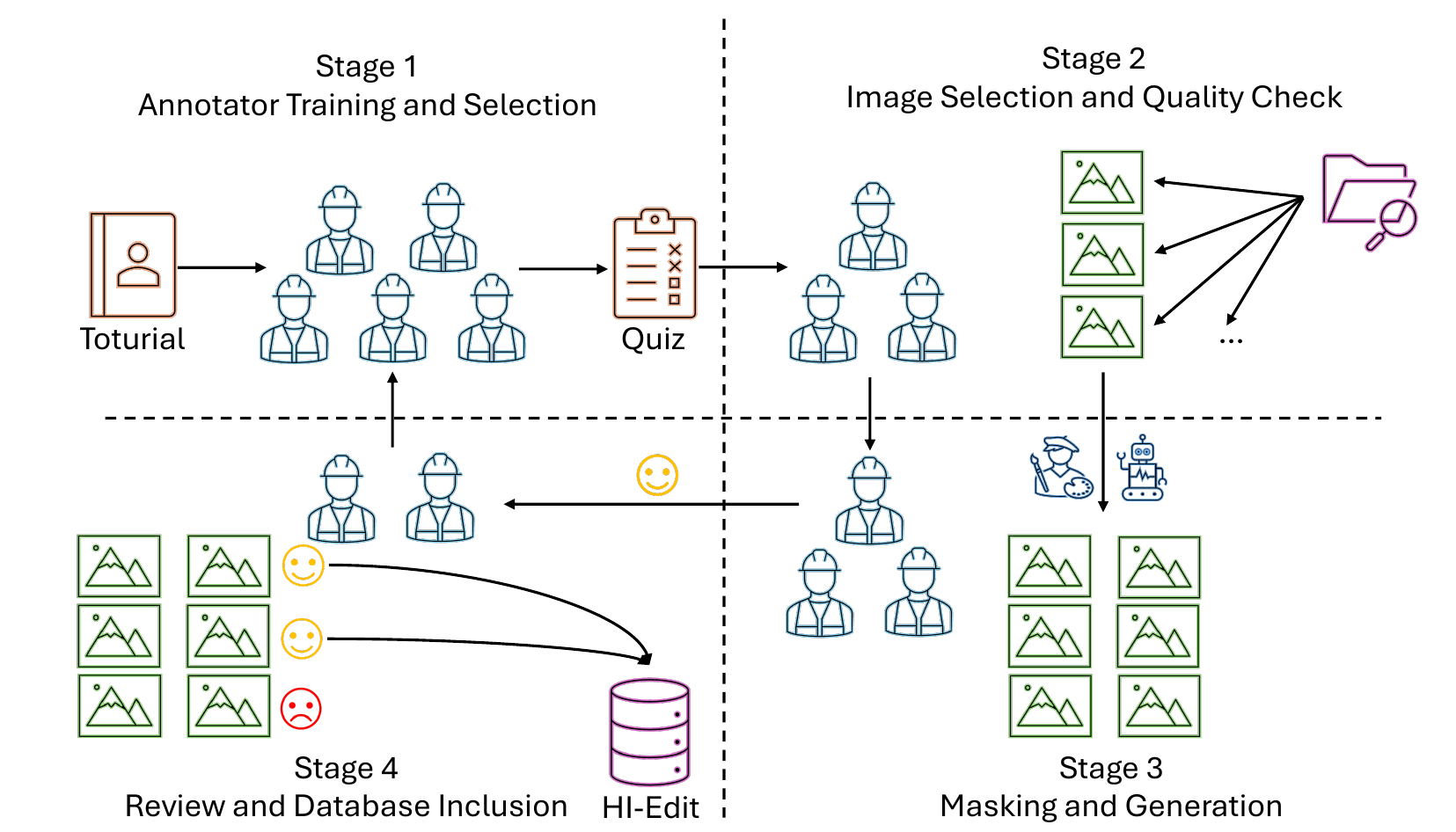}
\caption{Overview of data collection process.}
\label{fig:pipeline}
\end{figure}
In the \textbf{first stage}, we design a comprehensive tutorial and quiz to ensure high-quality annotations. The tutorial provides detailed guidance on effectively using the DALL-E 2 platform, along with essential annotation guidelines. More information about the tutorial can be found in Appendix~\ref{app:book}. We recruit over ten workers from an internal platform, train them using the tutorial, and conduct a quiz to evaluate their understanding. The top ten performers are selected as annotators.

In the \textbf{second stage}, we carefully curate high-resolution images from Unsplash~\citep{ali2023unsplash} and assign them to the selected annotators. Each annotator assesses the assigned images for their suitability based on predefined quality criteria. Images that fail to meet these criteria are replaced with new candidates, while suitable images proceed to the next stage.

In the \textbf{third stage}, annotators create novel and diverse editing instructions for the curated images. They utilize the DALL-E 2 platform to define mask areas, generate edited images, and provide captions for the results. Each submission package includes the original image, the mask, the edited image, the editing instruction, and the caption. These submissions are then forwarded to administrators for quality review to make sure being aligned with human performance. 

In the \textbf{fourth stage}, administrators perform a two-tier quality review and human feedback. If the edited image meets the required quality standards but the accompanying instructions or captions are problematic, the submission is returned to stage three for re-annotation. Submissions with poor editing quality are discarded. We refer to this process as \textbf{human-rewarded}, as annotators with good performance receive higher rewards, while those with poor performance are removed from the annotator teams. Examples of failure cases excluded from \name can be found in Appendix~\ref{app:failure}. Data pairs that pass the quality threshold are included in the final \name dataset. Over the course of the annotation process, approximately 20,000 images were annotated, with 5,751 high-quality images retained in the final dataset.

Finally, we leverage Llama 3.2-Vision~\citep{dubey2024llama} to generate refined captions for all original and edited images, ensuring consistency and clarity across the dataset.

\begin{figure}[]
	\centering  
	\vspace{-1.5mm}
	\includegraphics[width=13.5cm]{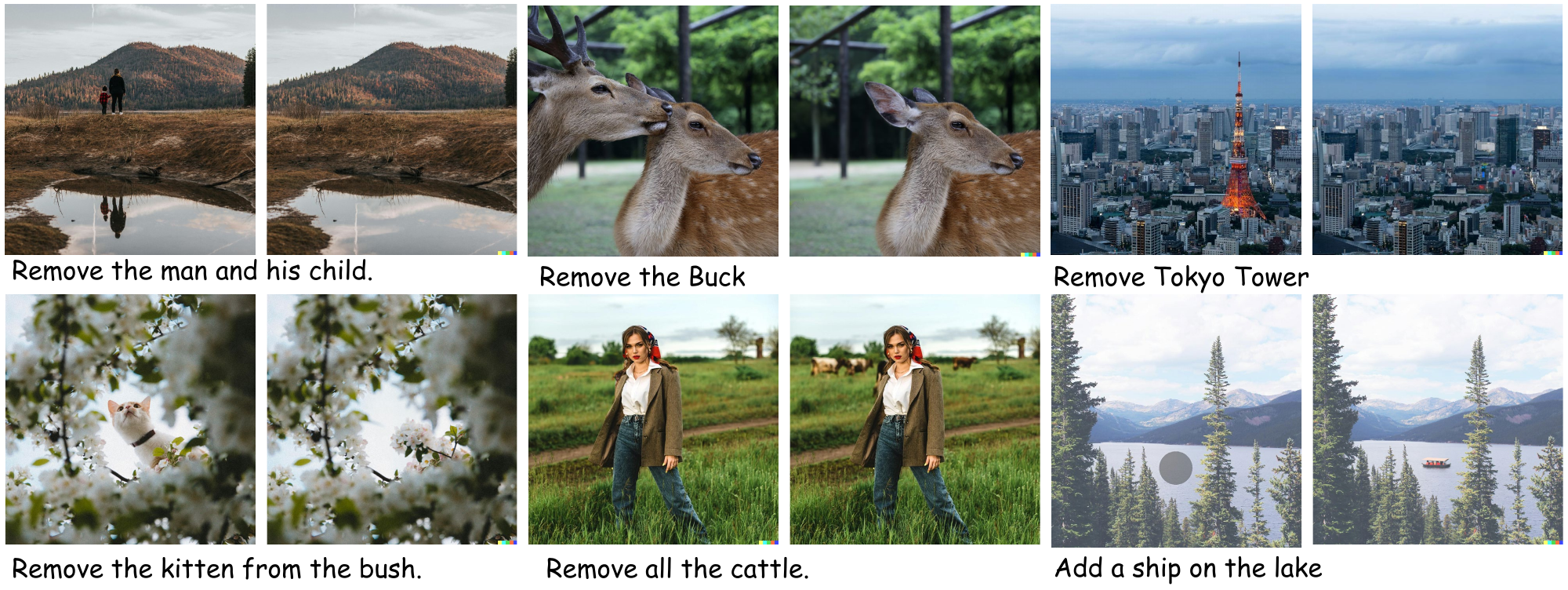}
        \includegraphics[width=13.5cm]{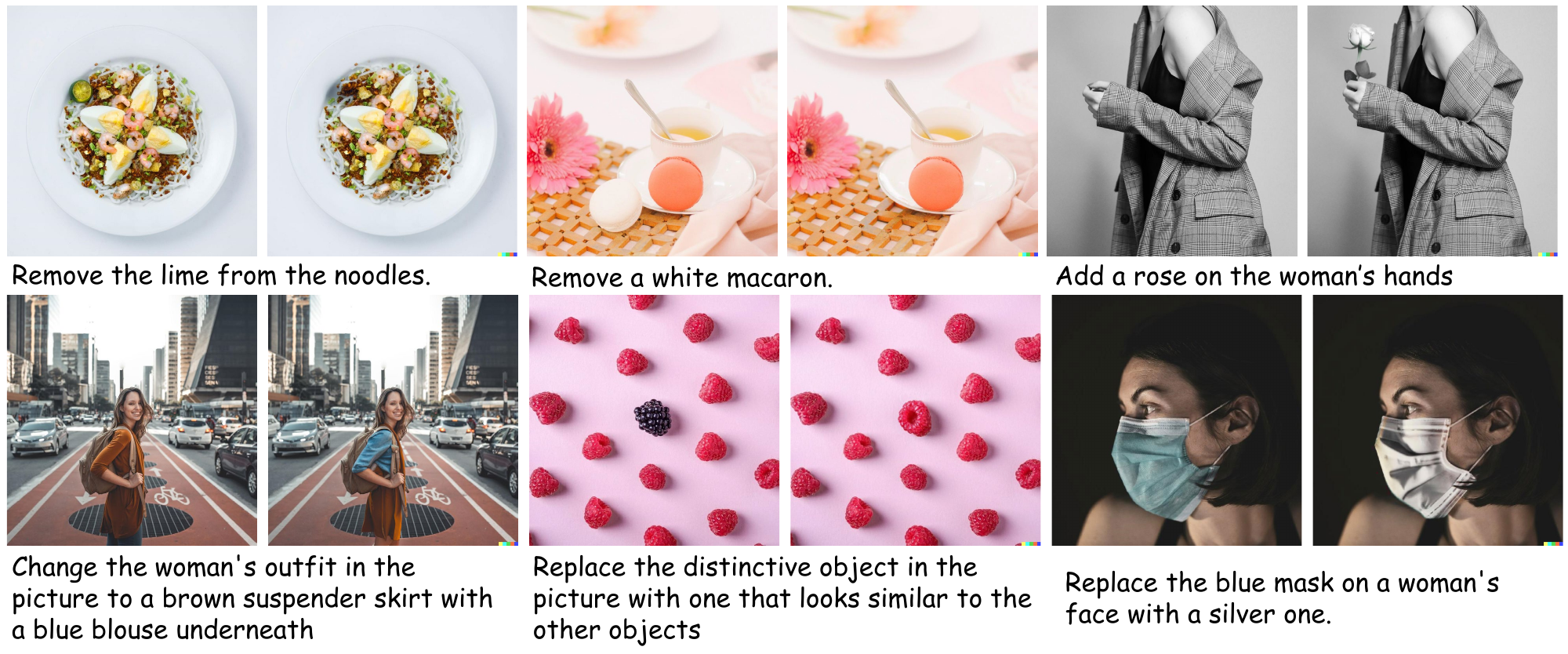}
        \includegraphics[width=13.5cm]{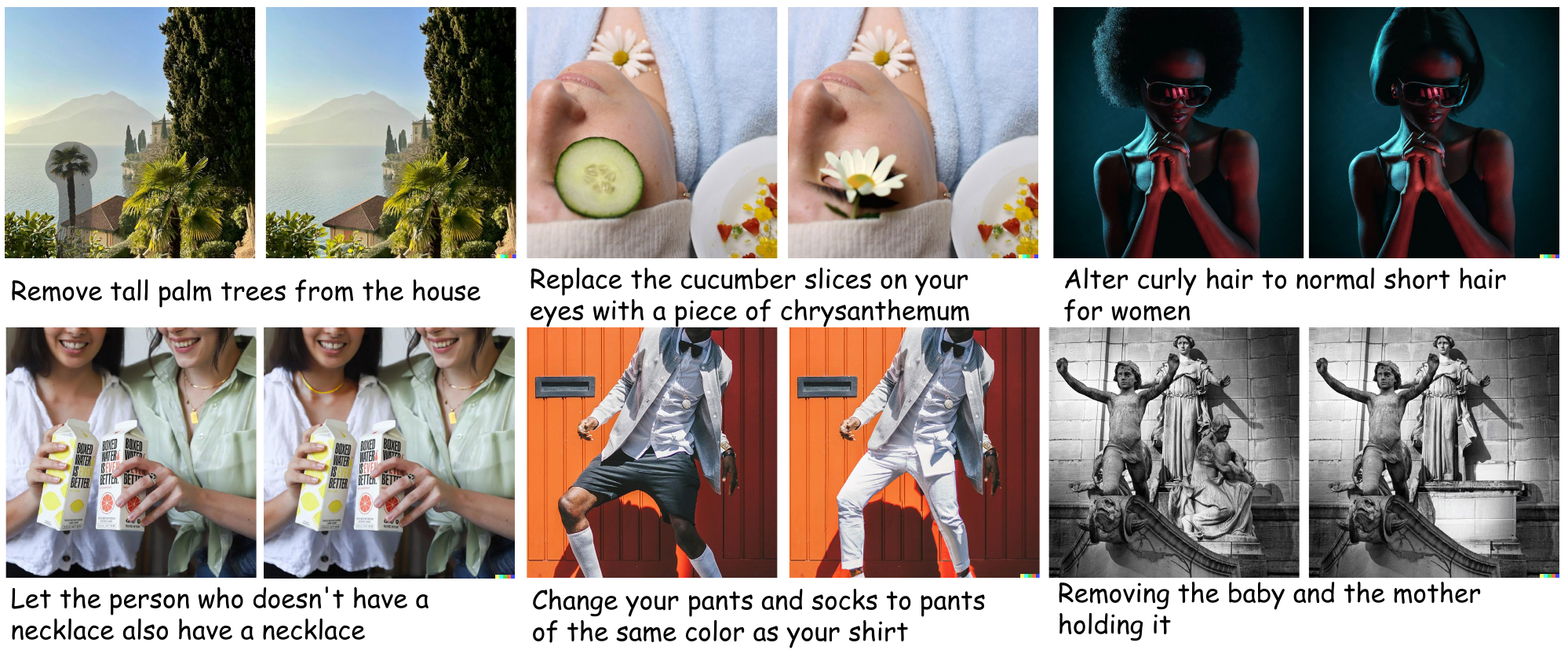}
        \includegraphics[width=13.5cm]{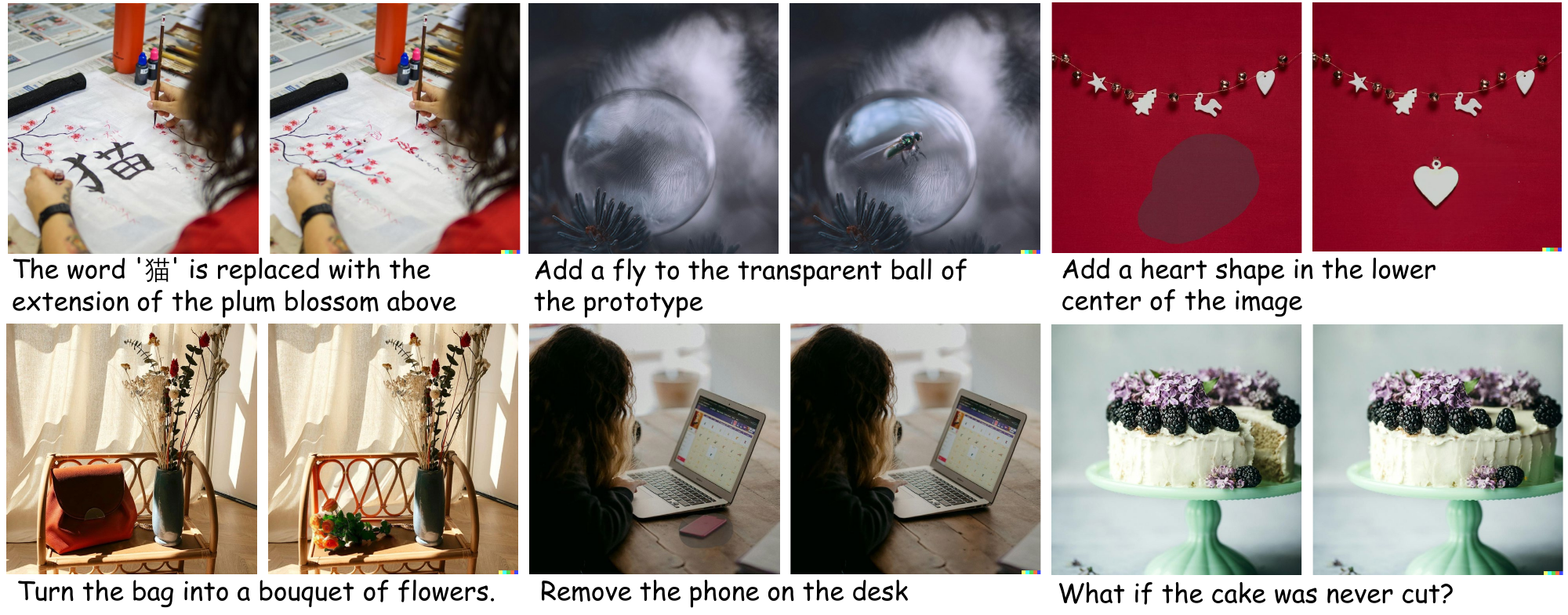}
	\vspace{-1.5mm}
    	\caption{
                More examples of instruction-guided image editing in \name{}.
            }
	\label{fig:more_ex}  
	\vspace{-3mm}
\end{figure}

\begin{table}[h]
    \small
    \centering
    \caption{Comparison of existing image editing datasets. ``Real Image for Edit'' denotes whether real images are used for editing instead of images generated by models. ``Real-world Scenario'' indicates whether images edited by users in the real world are included. ``Human'' denotes whether human annotators are involved. ``Ability Classification'' refers to evaluating the edit ability in different dimensions. ``Mask'' indicates whether rendering masks for editing is supported. ``Non-Mask Editing'' denotes the ability to edit without mask input.}
    \label{tab:comparison}
    \vspace{-1mm}
    \resizebox{1.0\columnwidth}{!}{
        \begin{tabular}{c|cccccc}
        \toprule
            Dataset & Real Image for Edit & Real-world Scenario & Human & Ability Classification & Mask & Non-Mask Editing \\
        \toprule
        InstructPix2Pix~\citep{brooks2023instructpix2pix} & \crossmark & \crossmark & \crossmark & \crossmark & \crossmark & \checkmark \\
        MagicBrush~\citep{zhang2024magicbrush} & \checkmark & \crossmark & \checkmark & \crossmark & \checkmark & \crossmark \\
        GIER~\citep{shi2020benchmark} & \checkmark & \checkmark & \checkmark & \crossmark & \crossmark & \crossmark \\
        MA5k-Req~\citep{shi2021learning} & \checkmark & \checkmark & \checkmark & \crossmark & \crossmark & \crossmark \\
        TEdBench~\citep{kawar2023imagic} & \checkmark & \checkmark & \checkmark & \crossmark & \crossmark & \checkmark \\
        HQ-Edit~\citep{hui2024hq} & \crossmark & \crossmark & \crossmark & \crossmark & \crossmark & \checkmark \\
        SEED-Data-Edit~\citep{ge2024seed} & \checkmark & \checkmark & \checkmark & \crossmark & \crossmark & \checkmark \\
        AnyEdit~\citep{yu2024anyeditmasteringunifiedhighquality} & \checkmark & \checkmark & \crossmark & \crossmark & \checkmark & \checkmark \\
        \name{} & \checkmark & \checkmark & \checkmark & \textbf{6} & \checkmark & \checkmark \\
        \toprule
        \end{tabular}
    }
    \vspace{-3mm}
\end{table}

\section{Dataset Statistics}\label{sec:stat}
\textbf{Related Datasets Comparison}.
InstructPix2Pix~\citep{brooks2023instructpix2pix} utilizes Prompt-to-Prompt~\citep{hertz2022prompt} to generate source and target images based on input and edit prompts from the LAION-Aesthetics~\citep{schuhmann2022laion} dataset. However, all images are model-generated, thereby lacking real-world authenticity.
MagicBrush~\citep{zhang2024magicbrush} employs crowdworkers on Amazon Mechanical Turk (AMT) to manually annotate images from the MS COCO dataset, using the DALL-E 2 platform for multi-round editing annotations. Although it offers diversity, all images are sourced from MS COCO and only support masked editing.
HQ-Edit leverages GPT-4~\citep{openai2023gpt} to generate image descriptions and editing instructions, creating paired images with GPT-4V~\citep{openai2023gptv} and DALL-E 3~\citep{betker2023improving}. These paired images are divided into source and target images, with instructions rewritten by GPT-4V. Nonetheless, this method often fails to preserve the fine-grained details of the source image in the target image, resulting in generated images that lack realism.
GIER~\citep{shi2020benchmark} and MA5k-Req~\citep{shi2021learning} only support filter changes, offering very limited richness.
SEED-Data-Edit~\citep{ge2024seed} boasts a larger dataset and supports unmasked editing, but it lacks capability classification and does not provide masks.
A more detailed comparison can be found in Table~\ref{tab:comparison}. Additionally, although SEED-Data-Edit has a large scale, the annotation process involves using several VLMs to generate instructions and captions, which may introduce hallucinations~\citep{yu2024hallucidoctor, liu2024survey}. In contrast, \name{} uses high-resolution original images, selecting higher-quality images as sources through VLM scoring and human selection. It supports both masked and unmasked editing, providing a high-quality dataset and benchmark for image editing.

\textbf{High Image Resolution}. The distribution of input image resolutions is depicted in Figure~\ref{fig:pie}(c). Most images in our dataset (62.3\%) have a resolution greater than 1000, with 33.8\% of them exceeding 1200. For image edited by DALL-E2~\citep{ramesh2022hierarchical}, lower resolution images are upsampled first, meaning that higher input image resolutions result in higher fidelity in the edited output images, as output images with a fixed size of $1024\times 1024$. In contrast, MagicBrush has only 46.6 input images above 1000, 25.3\% less than us, with the rest of input images has only $512\times 512$ resolution.

\textbf{Support Non-mask Editing}. \name~operates on real images and does not require additional inputs such as image masks or extra views of the object. As shown in Figure~\ref{fig:pie}(a), \name~provides masks for all data, with 46.5\% of the data supporting editing without masks. In contrast, datasets like MagicBrush require masks for editing. We believe this feature makes \name~more versatile and applicable, as real-world editing often does not involve using masks.

\textbf{Diverse Data Sources}. As illustrated in Figure~\ref{fig:pie}(b), the majority of our data originates from Unsplash~\citep{ali2023unsplash}, a website dedicated to photography. The images on this platform are known for their exceptional aesthetic quality, characterized by professional composition, lighting, and subject matter~\citep{li2023variational}. Our dataset is a carefully curated subset, selected from a pool of 57,000 crawled images, ensuring high quality and rich diversity.

\begin{figure}[h!]
    \centering  
    \includegraphics[width=0.99\textwidth]{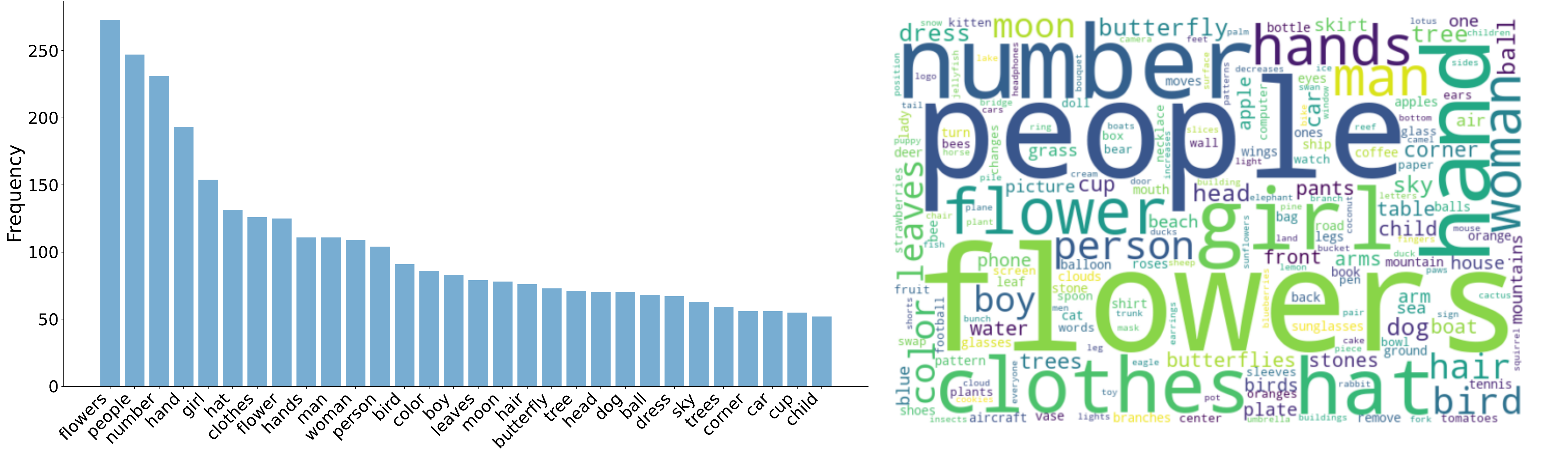}
    \vspace{-1mm}
    \caption{
(a) The distribution chart of the first 30 objects in the editing instructions for \name.
(b) The word cloud representation of the objects present in the editing instructions for \name.
}
    \label{fig:stat01}  
    \vspace{-2mm}
\end{figure}

\textbf{Rich Editing Instruction}. \name~encompasses a diverse array of edit instructions, including object addition, replacement, and removal, action changes, color alterations, text or pattern modifications, and object quantity adjustments. Keywords associated with each edit type span a wide spectrum, encompassing various objects, actions, and attributes, as depicted in Figure~\ref{fig:stat:app:1} and Figure~\ref{fig:stat:app:2}. This diversity underscores \name's ability to capture a comprehensive range of editing scenarios, facilitating robust training and evaluation of instruction-guided image editing models.

\begin{figure}[h!]
	\centering  
	\includegraphics[width=1.0\textwidth]{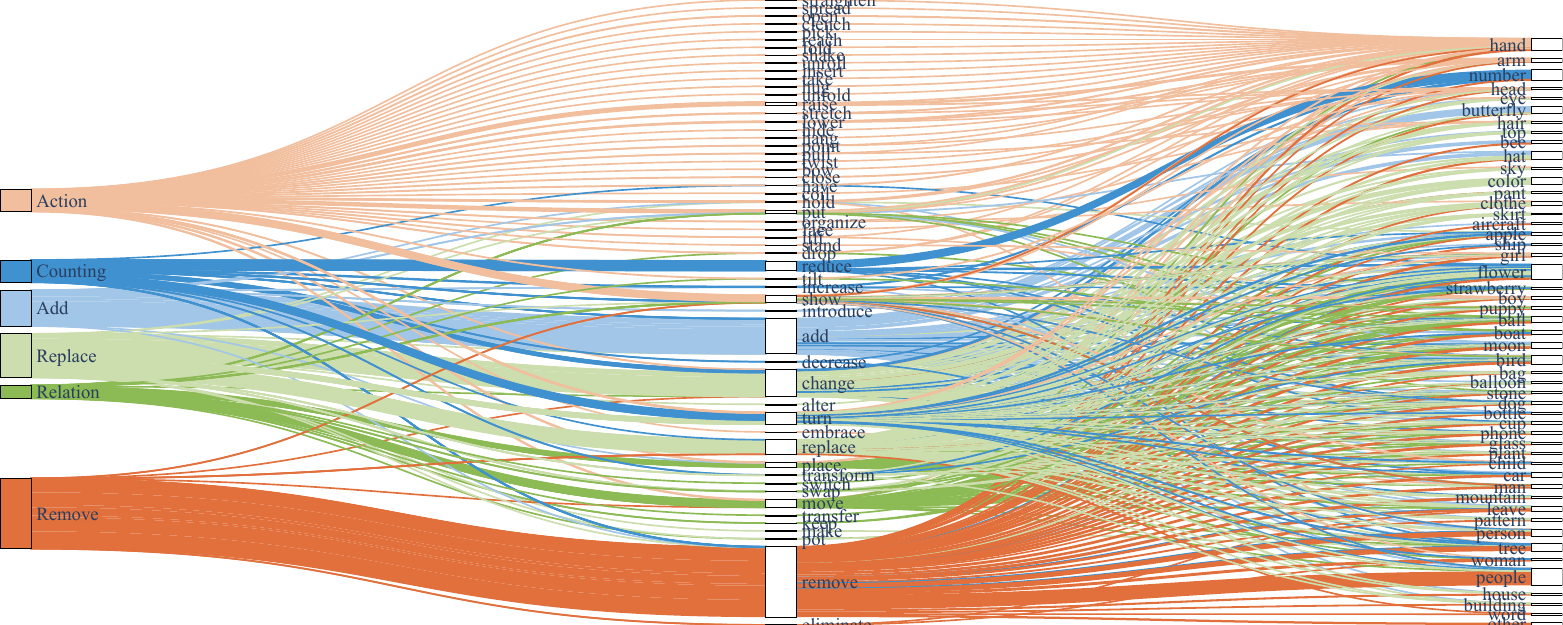}
	\caption{\textbf{The river chart of \name-full}. The first node of the river represents the type of edit, the second node corresponds to the verb extracted from the instruction, and the final node corresponds to the noun in the instruction. To maintain clarity, we only selected the top 50 most frequent nouns. The  river chart of \name-core can be seen in Figure~\ref{fig:flow_core} in Appendix.}
	\label{fig:flow_full}  
\end{figure}

\begin{figure}[h!]
	\centering  
	\includegraphics[width=0.99\textwidth]{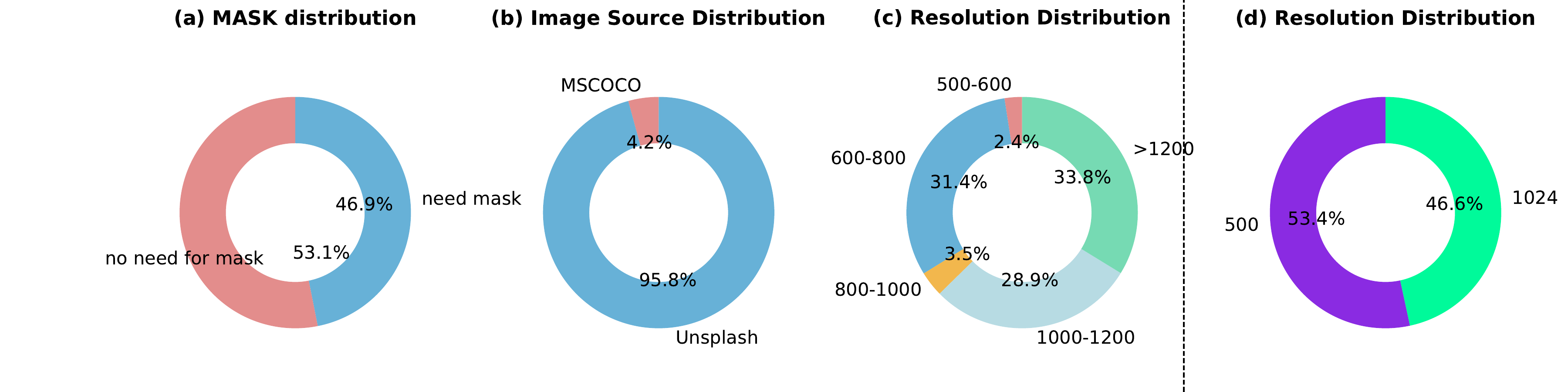}
	\caption{
	(a) The distribution of images for which \name{} requires masking, where \textit{no need for mask} refers to editing instructions that are already clear and comprehensive enough, and we believe that no masking is necessary for the model to complete the editing.
        (b) The distribution of the sources of all input images for \name.
        (c) The distribution of resolutions for all input images in \name.
        (d) The distribution of resolutions for all input images in MagicBrush.
        }
	\label{fig:pie}  
\end{figure}

\section{HI-EDIT Benchmark}\label{benchmark}

\textbf{Baselines.}
To provide the performance benchmark on \name{}, we consider multiple baselines in both mask-free and mask-provided settings. For all baselines, we adopt the default hyperparameters available in the official code repositories to guarantee reproducibility and fairness. 

For \textbf{mask-free baselines}, we consider:

\begin{itemize}
    \item \textit{InstructPix2Pix}~\citep{brooks2023instructpix2pix}. InstructPix2Pix Utilizes automatically generated instruction-based image editing data by large language models to fine-tune Stable Diffusion~\citep{rombach2022high}, enabling instruction-based image editing during inference without requiring any test-time tuning.
    \item \textit{MGIE}~\citep{fu2023guiding}. MLLM-Guided Image Editing (MGIE) explores how Multimodal Large Language Models~\citep{chow2024unified, pan2024auto, li2023fine} assist in generating edit instructions. MGIE learns to derive expressive instructions and provides explicit guidance for the editing process. The model integrates this visual imagination and performs image manipulation through end-to-end training.
    \item \textit{HIVE SD1.5}~\citep{zhang2024hive}. HIVE stands for \textbf{H}uman Feedback for \textbf{I}nstructional \textbf{V}isual \textbf{E}diting. The reward model is trained on supplementary data annotated by humans who rank the variant outputs of the fine-tuned InstructPix2Pix model. HIVE undergoes further fine-tuning using this reward model derived from these human rankings.
    \item \textit{MagicBrush}~\citep{zhang2024magicbrush}. MagicBrush curates a well-structured editing dataset with detailed human annotations and fine-tunes its model on this dataset using the InstructPix2Pix~\cite{brooks2023instructpix2pix} framework.
\end{itemize}

For \textbf{mask-provided baselines}, we consider:

\begin{itemize}
    \item \textit{Blended Latent Diffusion SDXL}~\citep{avrahami2023blended}. Latent Diffusion~\citep{rombach2022high} can generate an image from a given text (text-to-image LDM). However, it lacks the ability to edit an existing image in a local way. Blended Latent Diffusion incorporates Blended Diffusion~\citep{avrahami2022blended} into text-to-image LDM by utilizing CLIP~\citep{radford2021learning} guidance during the masked region denoising process and integrates it with the context from the noisy source image at each denoising timestep to enhance the region-context consistency of the generated target image.
    \item \textit{GLIDE}~\citep{nichol2021glide}. GLIDE stands for \textbf{G}uided
    \textbf{L}anguage to \textbf{I}mage \textbf{D}iffusion for Generation and \textbf{E}diting. To achieve better results on image editing tasks, OpenAI fine-tunes their model by modifying the model architecture to have four additional input channels: a second set of RGB channels and a mask channel. In addition, they initialize the corresponding input weights for these new channels to zero before fine-tuning. During fine-tuning, random regions of training examples are erased, and the remaining portions are fed into the model along with a mask channel as additional conditioning information.
    \item \textit{aMUSEd}~\citep{patil2024amused}. aMUSEd is a lightweight text-to-image model based on the MUSE architecture, which supports zero-shot image editing. For editing tasks, the mask directly determines which tokens are initially masked.
    \item \textit{Meissonic}~\citep{bai2024meissonic}. Meissonic is a non-autoregressive mask image modeling text-to-image synthesis model that can generate 1024 x 1024 high-resolution images. By incorporating a comprehensive suite of architectural innovations, advanced positional encoding strategies, and optimized sampling conditions, Meissonic substantially improves MIM's performance and efficiency to a level comparable with state-of-the-art diffusion models like SDXL. Due to the architecture of masked generative transformer, Meissonic also supports zero-shot image editing by masking the corresponding tokens.
\end{itemize}

\noindent
\textbf{Evaluation Metrics.}
Follow the similar settings from previous works~\citep{brooks2023instructpix2pix, zhang2024magicbrush}, we utilize L1 and L2 to measure the average pixel-level absolute difference between the generated image and ground truth image, and CLIP-I and DINO to measure the image quality with the cosine similarity between the generated image and reference ground truth image using their CLIP~\citep{radford2021learning} and DINO~\citep{caron2021emerging} embeddings,
and CLIP-T~\citep{ruiz2023dreambooth, chen2024subject} to measure the text-image alignment with the cosine similarity between local descriptions and generated images CLIP embeddings.

\begin{table}[t]
\small
\centering
\caption{Quantitative study on mask-free baselines on \name{}. The best results are marked in \textbf{bold}.
}
\begin{tabular}{lccccc}
\toprule
 \multicolumn{1}{c}{\textbf{Methods}} & L1$\downarrow$  & L2$\downarrow$  & CLIP-I$\uparrow$ & DINO$\uparrow$  & CLIP-T$\uparrow$ \\ \midrule
\multicolumn{6}{c}{\textit{\name{}-full}}                           \\ \cmidrule{1-6} 
 InstructPix2Pix~\citep{brooks2023instructpix2pix}           & 0.1601 &0.0551 &0.7716 &0.5335 &0.2591                        \\
 MGIE~\citep{fu2023guiding}          & 0.1240 &0.0535 &0.8697 &0.7221 &0.2661     \\
 HIVE SD1.5~\citep{zhang2024hive}          & 0.1014                             & \textbf{0.0278    }                         & 0.8526                               & 0.7726                             & \textbf{0.2777 }                              \\
 MagicBrush~\citep{zhang2024magicbrush}          & \textbf{0.0807 }  &0.0298 &\textbf{0.8915 }   &\textbf{0.7963} &0.2676    \\ \cmidrule{1-6} 
 \multicolumn{6}{c}{\textit{\name{}-core}}                                                  \\ \cmidrule{1-6} 
 InstructPix2Pix~\citep{brooks2023instructpix2pix}           &0.1625    &0.0570    &0.7627    &0.5349    &0.2533      \\
 MGIE~\citep{fu2023guiding}          & 0.1294    &0.0610    &0.8670    &0.7359    &0.2589   \\
 HIVE SD1.5~\citep{zhang2024hive}          &0.1162    &0.0373    &0.8441    &0.7038    &0.2563     \\
 MagicBrush~\citep{zhang2024magicbrush}          & \textbf{0.0760 }   &\textbf{0.0283   } &\textbf{0.8946}    &\textbf{0.8121 }   &\textbf{0.2619}     \\ \bottomrule

\end{tabular}
\label{tab:mask-free-quantitative}
\end{table}
\begin{table}[h]
\small
\centering
\caption{
Quantitative study on mask-provided baselines on \name{}. The best results are marked in \textbf{bold}.
}
\begin{tabular}{lccccc}
\toprule
 \multicolumn{1}{c}{\textbf{Methods}} & L1$\downarrow$  & L2$\downarrow$  & CLIP-I$\uparrow$ & DINO$\uparrow$  & CLIP-T$\uparrow$ \\ \midrule
\multicolumn{6}{c}{\textit{\name{}-full}}                           \\ \cmidrule{1-6} 
 Blended Latent Diff. SDXL~\citep{avrahami2023blended}            &0.0481    &0.0151    &0.9178    &0.8481    &0.2681    \\
 GLIDE~\citep{nichol2021glide}       &\textbf{0.0391}    &\textbf{0.0120}    &\textbf{0.9388}    &0.8800    &0.2676    \\
 aMUSEd~\citep{patil2024amused}   &0.0673    &0.0187    &0.9149    &0.8588    &\textbf{0.2771}  \\
 Meissonic~\citep{bai2024meissonic}   &0.0627    &0.0177    &0.9324    &\textbf{0.8806}    &0.2710      \\ \cmidrule{1-6} 
 \multicolumn{6}{c}{\textit{\name{}-core}}                                                  \\ \cmidrule{1-6} 
 Blended Latent Diff. SDXL~\citep{avrahami2023blended} &0.0496    &0.0162    &0.9116    &0.8550    &0.2640   \\
 GLIDE~\citep{nichol2021glide}   &\textbf{0.0379}    &\textbf{0.0113 }   &\textbf{0.9413}    &\textbf{0.8961}    &0.2656   \\
 aMUSEd~\citep{patil2024amused}   &0.0665    &0.0184    &0.9138    &0.8743    &\textbf{0.2747}   \\
 Meissonic~\citep{bai2024meissonic}    &0.0608    &0.0166    &0.9348    &0.8943    &0.2694    \\ \bottomrule
  \multicolumn{6}{c}{\textit{\name{}-mask}}                                                  \\ \cmidrule{1-6} 
 Blended Latent Diff. SDXL~\citep{avrahami2023blended}   &0.0478    &0.0154    &0.9065    &0.8223    &0.2650     \\
 GLIDE~\citep{nichol2021glide}    &\textbf{0.0377}    &\textbf{0.0117 }   &\textbf{0.9343 }   &0.8687    &0.2665   \\
 aMUSEd~\citep{patil2024amused}    &0.0654    &0.0179    &0.9097    &0.8497    &\textbf{0.2785}   \\
 Meissonic~\citep{bai2024meissonic}   &0.0604    &0.0166    &0.9303    &\textbf{0.8783}    &0.2715    \\ \bottomrule

\end{tabular}
\label{tab:mask-provided-quantitative}
\end{table}

\noindent
\textbf{\name{} Benchmark.}
Tables~\ref{tab:mask-free-quantitative} and~\ref{tab:mask-provided-quantitative} summarize the quantitative results for mask-free and mask-provided methods, respectively. Mask-free methods are given only textual instructions to edit images, while mask-provided methods receive both instructions and corresponding masks.

\begin{table}[!htb]
\small
\centering
\caption{Quantitative study on six different types of editing instructions on \name{}. The best results are marked in \textbf{bold}.
}
\begin{tabular}{lccccc}
\toprule
 \multicolumn{1}{c}{\textbf{Methods}} & L1$\downarrow$  & L2$\downarrow$  & CLIP-I$\uparrow$ & DINO$\uparrow$  & CLIP-T$\uparrow$ \\ \midrule

 \multicolumn{6}{c}{\textit{\name{}-Add}}                                                  \\ \cmidrule{1-6} 
 InstructPix2Pix~\citep{brooks2023instructpix2pix}    &0.1152    &0.0329    &0.8135    &0.6230    &0.2764  \\
 MGIE~\citep{fu2023guiding}     &0.0934    &0.0274    &0.8770    &0.7391    &0.2806  \\
 HIVE SD1.5~\citep{zhang2024hive}    &0.0885    &0.0234    &0.8863    &0.7811    &0.2706   \\
 MagicBrush~\citep{zhang2024magicbrush}    &0.0580    &0.0167    &0.9102    &0.8562    &0.2745    \\ Blended Latent Diff. SDXL~\citep{avrahami2023blended}     &0.0344    &\textbf{0.0073 }   &0.9285    &0.8856    &0.2665  \\
 GLIDE~\citep{nichol2021glide}   &\textbf{0.0315}    &0.0078    &\textbf{0.9410}    &\textbf{0.8995 }   &0.2600  \\
 aMUSEd~\citep{patil2024amused}   &0.0581    &0.0130    &0.9148    &0.8672    &\textbf{0.2695}    \\
 Meissonic~\citep{bai2024meissonic}    &0.0544    &0.0129    &0.9303    &0.8787    &0.2669   \\ \cmidrule{1-6}  

\multicolumn{6}{c}{\textit{\name{}-Action}}                           \\ \cmidrule{1-6} 
 InstructPix2Pix~\citep{brooks2023instructpix2pix}            &0.1324    &0.0398    &0.7514    &0.5789    &0.2617                \\
 MGIE~\citep{fu2023guiding}         &0.0982    &0.0383    &0.8788    &0.7909    &0.2658     \\
 HIVE SD1.5~\citep{zhang2024hive}        &0.0972    &0.0280    &0.8592    &0.7613    &0.2640         \\
 MagicBrush~\citep{zhang2024magicbrush}         &0.0723    &0.0245    &0.9028    &0.8357    &0.2668   \\  Blended Latent Diff. SDXL~\citep{avrahami2023blended}   &0.0416    &\textbf{0.0109}    &0.9391    &0.9015    &0.2712  \\
 GLIDE~\citep{nichol2021glide}   &\textbf{0.0384}    &0.0114    &\textbf{0.9487}    &0.9018    &0.2683    \\
 aMUSEd~\citep{patil2024amused}   &0.0629    &0.0156    &0.9230    &0.8919    &\textbf{0.2732}    \\
 Meissonic~\citep{bai2024meissonic}    &0.0577    &0.0145    &0.9430    &\textbf{0.9126 }   &0.2677    \\\cmidrule{1-6}

 \multicolumn{6}{c}{\textit{\name{}-Counting}}                    \\ \cmidrule{1-6} 
 InstructPix2Pix~\citep{brooks2023instructpix2pix}        &0.1628    &0.0586    &0.8124    &0.5850    &0.2716 \\
 MGIE~\citep{fu2023guiding}         &0.1380    &0.0641    &0.8726    &0.6971    &0.2716 \\
 HIVE SD1.5~\citep{zhang2024hive}   &0.1211    &0.0442    &0.8826    &0.7431    &0.2705                 \\
 MagicBrush~\citep{zhang2024magicbrush}    &0.1058    &0.0434    &0.8677    &0.7103    &0.2707    \\ Blended Latent Diff. SDXL~\citep{avrahami2023blended}    &0.0527    &0.0180    &0.9334    &0.8892    &0.2766    \\
 GLIDE~\citep{nichol2021glide}    &\textbf{0.0392 }   &\textbf{0.0127 }   &\textbf{0.9523 }   &\textbf{0.9104 }   &0.2772    \\
 aMUSEd~\citep{patil2024amused}   &0.0699    &0.0213    &0.9270    &0.8816    &\textbf{0.2814 }  \\
 Meissonic~\citep{bai2024meissonic}    &0.0674    &0.0217    &0.9394    &0.8967    &0.2750  \\ \cmidrule{1-6}

 \multicolumn{6}{c}{\textit{\name{}-Remove}}                                                  \\ \cmidrule{1-6} 
 InstructPix2Pix~\citep{brooks2023instructpix2pix}     &0.1624    &0.0504    &0.7240    &0.4188    &0.2325  \\
 MGIE~\citep{fu2023guiding}    &0.1259    &0.0572    &0.8677    &0.7235    &0.2525    \\
 HIVE SD1.5~\citep{zhang2024hive}    &0.1179    &0.0375    &0.8362    &0.6562    &0.2474                          \\
 MagicBrush~\citep{zhang2024magicbrush}      &0.0690    &0.0232    &0.8985    &0.8249    &0.2572   \\ Blended Latent Diff. SDXL~\citep{avrahami2023blended}   &0.0451    &0.0133    &0.9055    &0.8322    &0.2608   \\
 GLIDE~\citep{nichol2021glide}   &\textbf{0.0313}    &\textbf{0.0072}    &\textbf{0.9493}    &\textbf{0.9119}    &0.2661    \\
 aMUSEd~\citep{patil2024amused}    &0.0621    &0.0156    &0.9148    &0.8702    &\textbf{0.2715 }   \\
 Meissonic~\citep{bai2024meissonic}     &0.0557    &0.0132    &0.9367    &0.9048    &0.2673  \\ \cmidrule{1-6}  

  \multicolumn{6}{c}{\textit{\name{}-Relation}}                          \\ \cmidrule{1-6} 
 InstructPix2Pix~\citep{brooks2023instructpix2pix}           &0.1741    &0.0647    &0.8069    &0.5851    &0.2828     \\
 MGIE~\citep{fu2023guiding}       &0.1420    &0.0656    &0.8762    &0.7061    &0.2768   \\
 HIVE SD1.5~\citep{zhang2024hive}    &0.1298    &0.0460    &0.8689    &0.7005    &0.2793              \\
 MagicBrush~\citep{zhang2024magicbrush}       &0.0884    &0.0334    &0.8985    &0.7865    &0.2823  \\ Blended Latent Diff. SDXL~\citep{avrahami2023blended}    &0.0628    &0.0213    &\textbf{0.9190 }   &\textbf{0.8174}    &0.2832    \\
 GLIDE~\citep{nichol2021glide}      &\textbf{0.0553 }   &\textbf{0.0192}    &0.9136    &0.7983    &0.2755   \\
 aMUSEd~\citep{patil2024amused}    &0.0809    &0.0267    &0.9076    &0.8095    &\textbf{0.2862}   \\
 Meissonic~\citep{bai2024meissonic}     &0.0825    &0.0283    &0.9171    &0.8142    &0.2768   \\ \cmidrule{1-6}  
 
 \multicolumn{6}{c}{\textit{\name{}-Replace}}                                                  \\ \cmidrule{1-6} 
 InstructPix2Pix~\citep{brooks2023instructpix2pix}        &0.1910    &0.0770    &0.7887    &0.5692    &0.2697      \\
 MGIE~\citep{fu2023guiding}        &0.1391    &0.0620    &0.8603    &0.6946    &0.2698    \\
 HIVE SD1.5~\citep{zhang2024hive}     &0.1265    &0.0443    &0.8582    &0.7087    &0.2726                         \\
 MagicBrush~\citep{zhang2024magicbrush}       &0.0984    &0.0409    &0.8757    &0.7513    &0.2716   \\ Blended Latent Diff. SDXL~\citep{avrahami2023blended}   &0.0567    &0.0206    &0.9095    &0.8096    &0.2683   \\
 GLIDE~\citep{nichol2021glide}    &\textbf{0.0495}    &\textbf{0.0188}    &0.9194    &0.8247    &0.2663   \\
 aMUSEd~\citep{patil2024amused}  &0.0761    &0.0237    &0.9072    &0.8259    &\textbf{0.2861}   \\
 Meissonic~\citep{bai2024meissonic}   &0.0710    &0.0227    &\textbf{0.9239 }   &\textbf{0.8462 }   &0.2762  \\ \bottomrule

\end{tabular}
\label{tab:6-types-quantitative}
\end{table}

\begin{figure}[!htb]
	\centering  
	\includegraphics[width=1.0\textwidth]{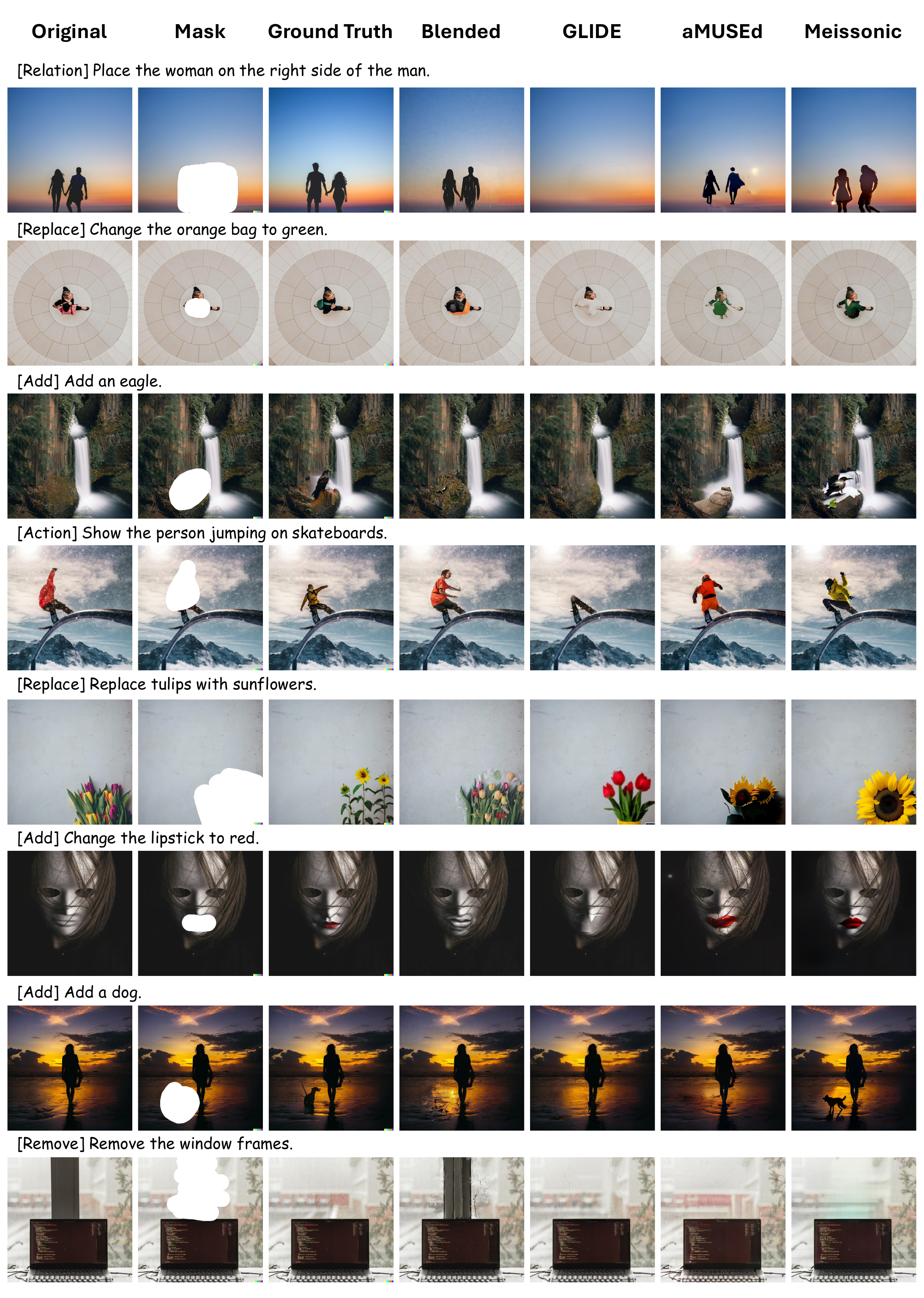}
	\caption{\textbf{Qualitative comparisons between mask-provided baselines.} The first three rows show the original images, corresponding masks, and ground truth edited images from DALL-E 2. The subsequent four rows present results generated by Blended Latent Diffusion SDXL, GLIDE, aMUSEd, and Meissonic, respectively.}

	\label{fig:vis_part1}  
\end{figure}

\begin{figure}[!htb]
	\centering  
	\includegraphics[width=1.0\textwidth]{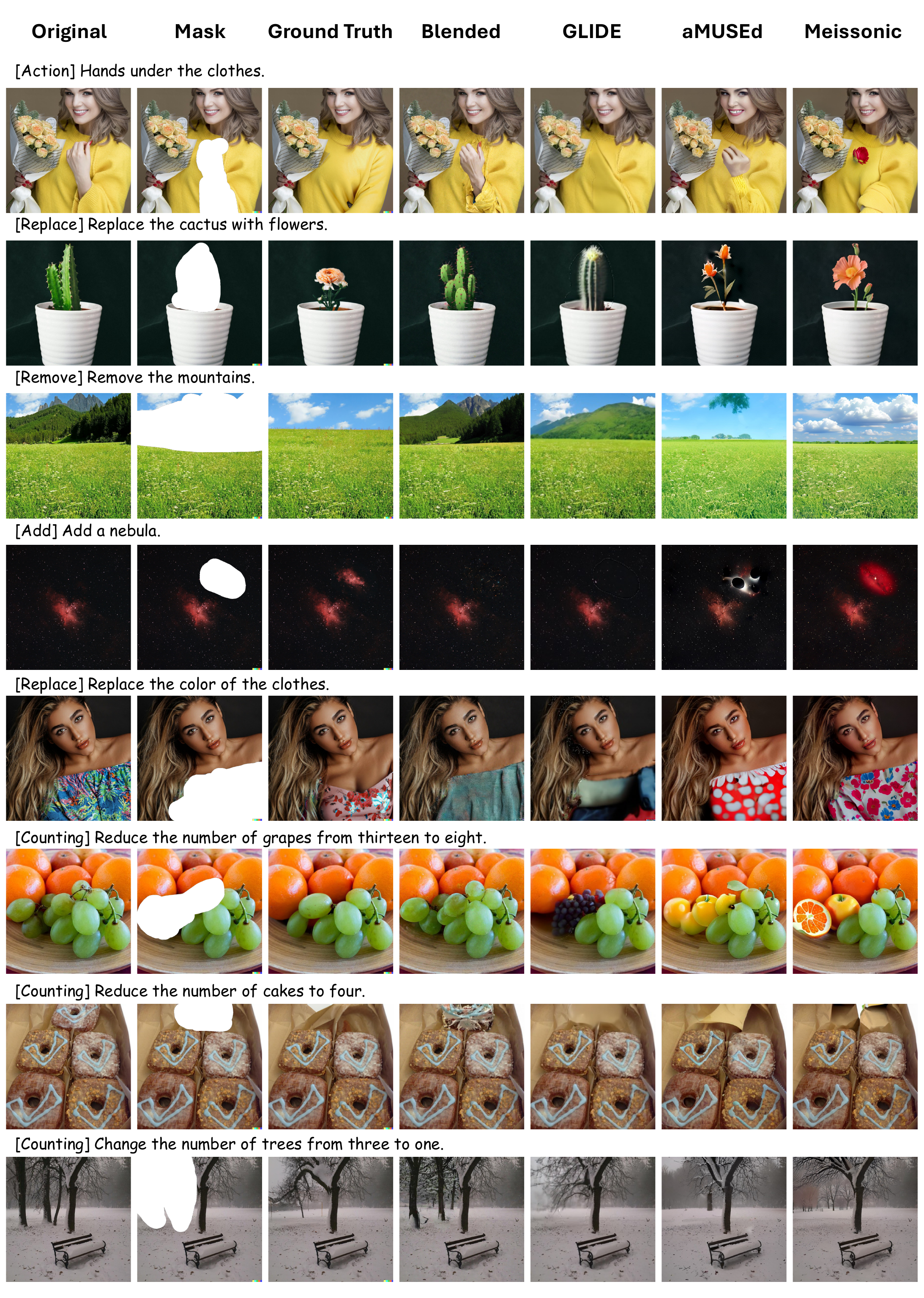}
	\caption{\textbf{Qualitative comparisons between mask-provided baselines.} The first three rows show the original images, corresponding masks, and ground truth edited images from DALL-E 2. The subsequent four rows present results generated by Blended Latent Diffusion SDXL, GLIDE, aMUSEd, and Meissonic, respectively.}

	\label{fig:vis_part2}  
\end{figure}
Additionally, Table~\ref{tab:6-types-quantitative} presents the quantitative results across six distinct types of editing instructions. We believe this categorization can facilitate fine-grained advancements in instruction-based image editing tasks. The table reveals several noteworthy observations: for instance, most methods perform better on \textit{Add} tasks than on \textit{Remove} tasks. Moreover, mask-provided methods generally achieve superior performance in semantic-level evaluation metrics compared to pixel-level ones.

To provide further insights, Figures~\ref{fig:vis_part1} and~\ref{fig:vis_part2} showcase visual examples of results from mask-provided methods. These examples highlight that existing methods perform well on \textit{Add} and \textit{Remove} editing tasks but struggle with more complex tasks such as \textit{Relation} and \textit{Action}. Furthermore, even for \textit{Add} tasks, challenges persist in cases requiring domain-specific knowledge or handling unfamiliar instructions, such as \textit{“Add a petal in the middle of the white puppy's forehead.”}

It is important to note that comparisons between methods might be unfair because of differences in implementation and fine-tuning. These tables are intended to establish a benchmark for \name{} to support future research and evaluation.

\section{Conclusion}

In this work, we introduce \name{}, a high-quality, human-rewarded dataset for instructional image editing. Previous large-scale editing datasets often incorporate minimal human feedback, leading to challenges in aligning datasets with human preferences. \name{} bridges this gap by employing human annotators to construct data pairs and administrators to provide feedback. Designed to address the growing demand for precise and versatile image editing capabilities, \name{} comprises six types of editing instructions: Action, Add, Counting, Relation, Remove, and Replace. The dataset stands out for its meticulous quality control, diverse sources, and inclusion of high-resolution images, offering unparalleled reliability and utility for model development. Furthermore, \name{} provides explicit differentiation between tasks requiring masks and those that do not, ensuring comprehensive support for a wide range of editing scenarios.

\section{Acknowledgements} 
This work was supported in part by NUS Start-up Grant A-0010106-00-00.

\clearpage

\clearpage 
\appendix
\section{More Figures} \label{app:figures}
\begin{figure}[h!]
	\centering  
	\includegraphics[width=0.86\textwidth]{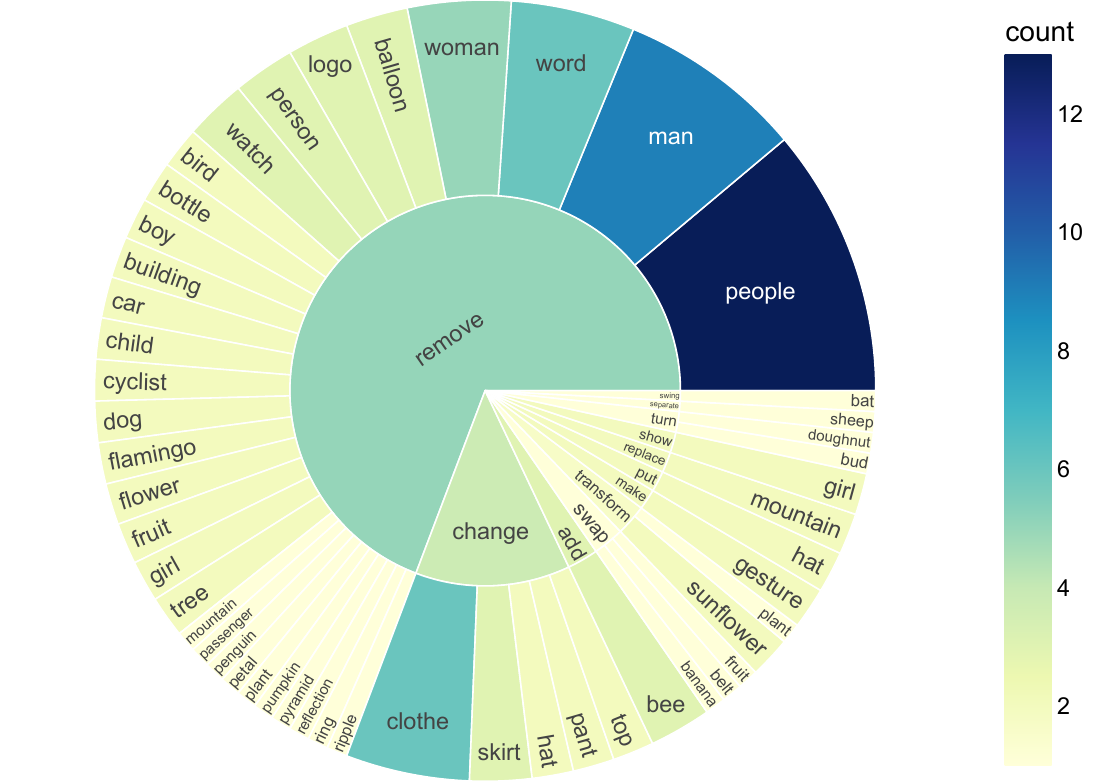}
	\vspace{-1mm}
            \caption{\textbf{An Overview of Keywords in \textbf{\name-core} Edit Instructions}: The inner circle represents the verb in the edit instruction, while the outer circle illustrates the noun following the verb in each instruction.}
	\label{fig:stat:app:1} 
\end{figure}

\begin{figure}[h!]
	\centering  
	\includegraphics[width=0.86\textwidth]{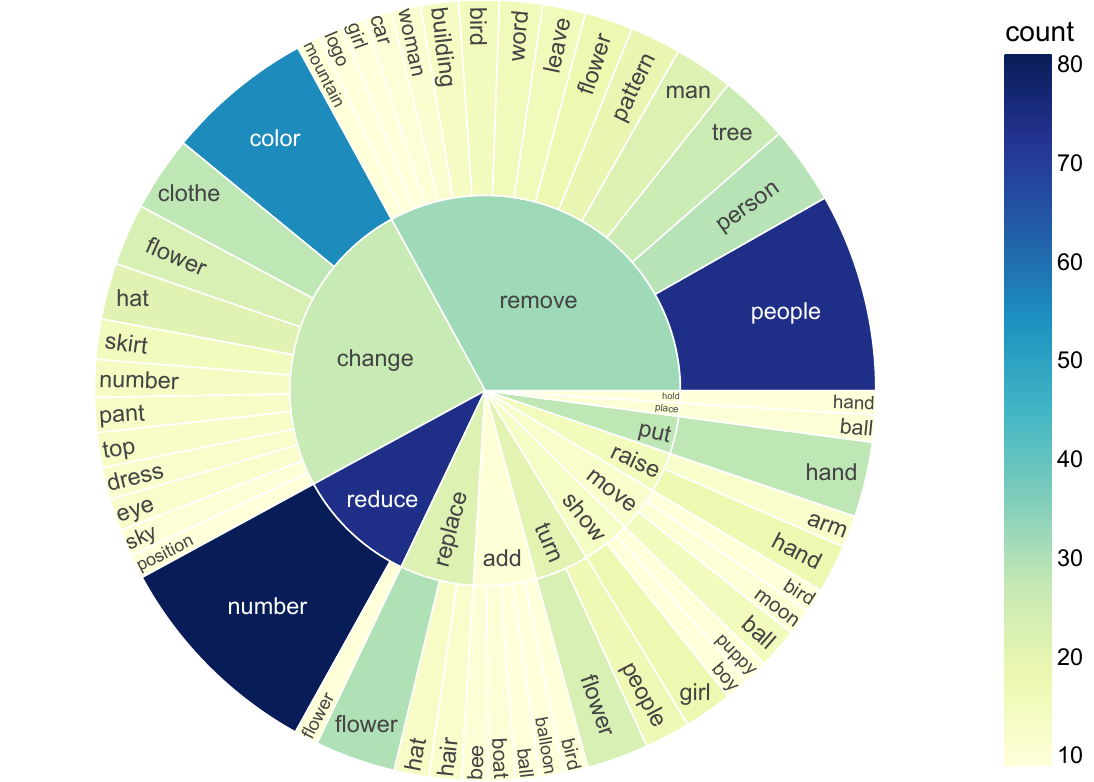}
	\vspace{-1mm}
    \caption{\textbf{An Overview of Keywords in \textbf{\name-full} Edit Instructions}: The inner circle represents the verb in the edit instruction, while the outer circle highlights the noun associated with the verb in each instruction.}
	\label{fig:stat:app:2} 
\end{figure}

In addition to the statistical charts mentioned in Section~\ref{sec:stat}, we also provide a sunburst chart analysis of the instructions, as shown in Figure~\ref{fig:stat:app:1} and Figure~\ref{fig:stat:app:2}. Due to space constraints, we have selected only the top 50 most frequent nouns for visualization.

\begin{figure}[h!]
	\centering  
	\includegraphics[width=1.0\textwidth]{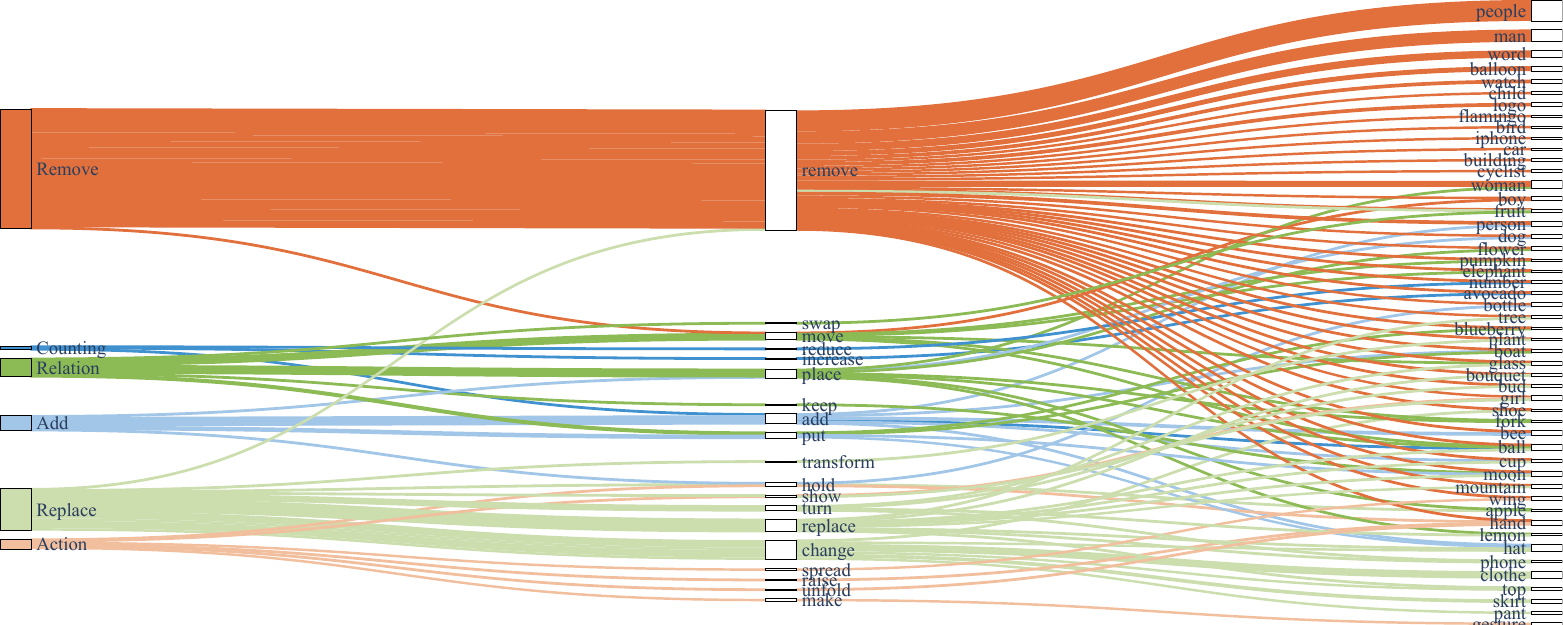}
	\vspace{-1mm}
	\caption{\textbf{The river chart of \name-core}. The first node of the river represents the type of edit, the second node corresponds to the verb extracted from the instruction, and the final node corresponds to the noun in the instruction. To maintain clarity, we only selected the top 50 most frequent nouns.}
	\label{fig:flow_core}  
\end{figure}

The river chart of \name-core is shown in Figure~\ref{fig:flow_core}. The full can be seen in Figure~\ref{fig:flow_full}.

\section{Guidance Book for Annotators}\label{app:book}
\subsection{Edit Cases for Annotators}
The following provides some annotation examples and the required submission content for annotators. We have conducted basic classification to help annotators develop a better understanding of the annotation task and to enrich the editing content as much as possible.

\textbf{(1) Object Related}. Object-centered editing can be categorized into the following four types.

\begin{figure}[h!]
	\centering  
	\includegraphics[width=1.0\textwidth]{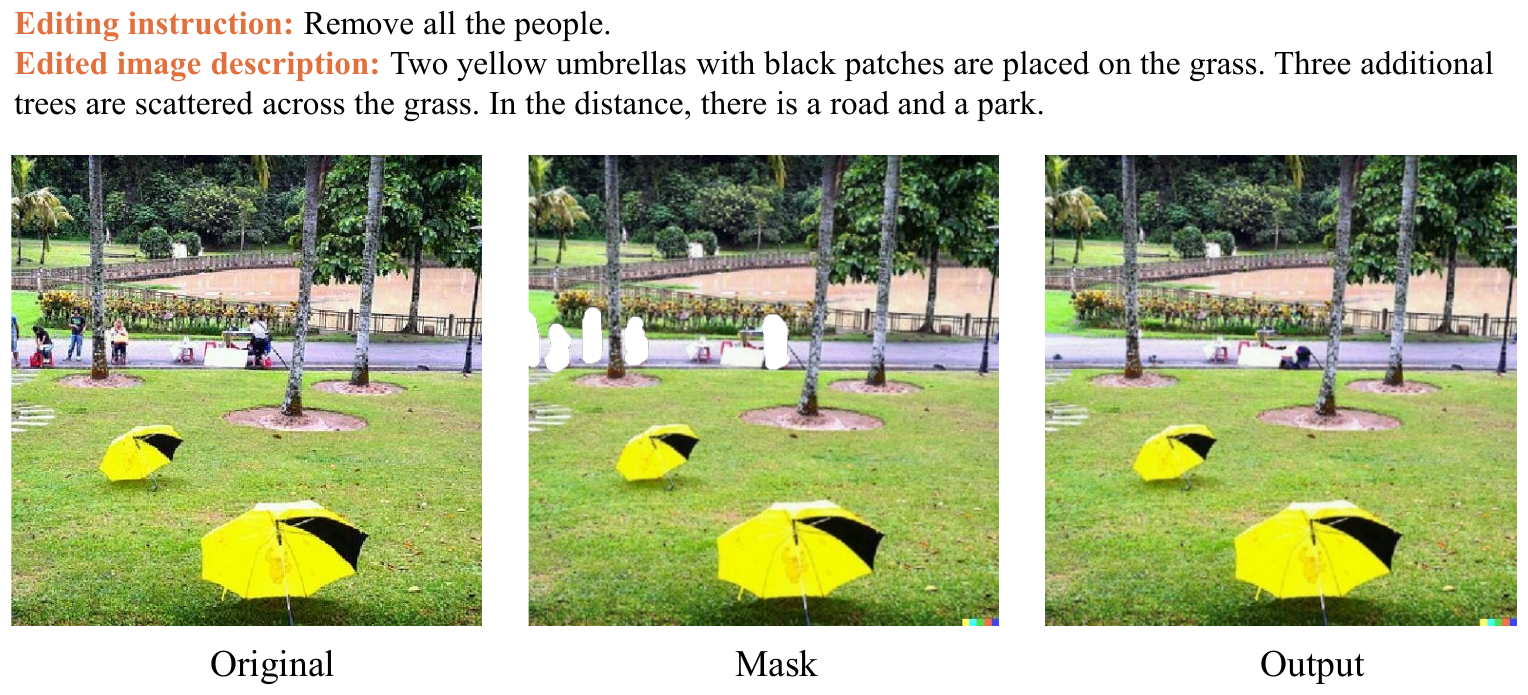}
	\vspace{-4mm}
	\caption{Case of Object Removal.}
	\label{fig:book1}  
\end{figure}

\textbf{(1.1) Object Removal}. As shown in Figure~\ref{fig:book1}, this task primarily involves removing certain objects from an image, typically those that are more prominent or easily distinguishable.

\begin{figure}[h!]
	\centering  
	\includegraphics[width=1.0\textwidth]{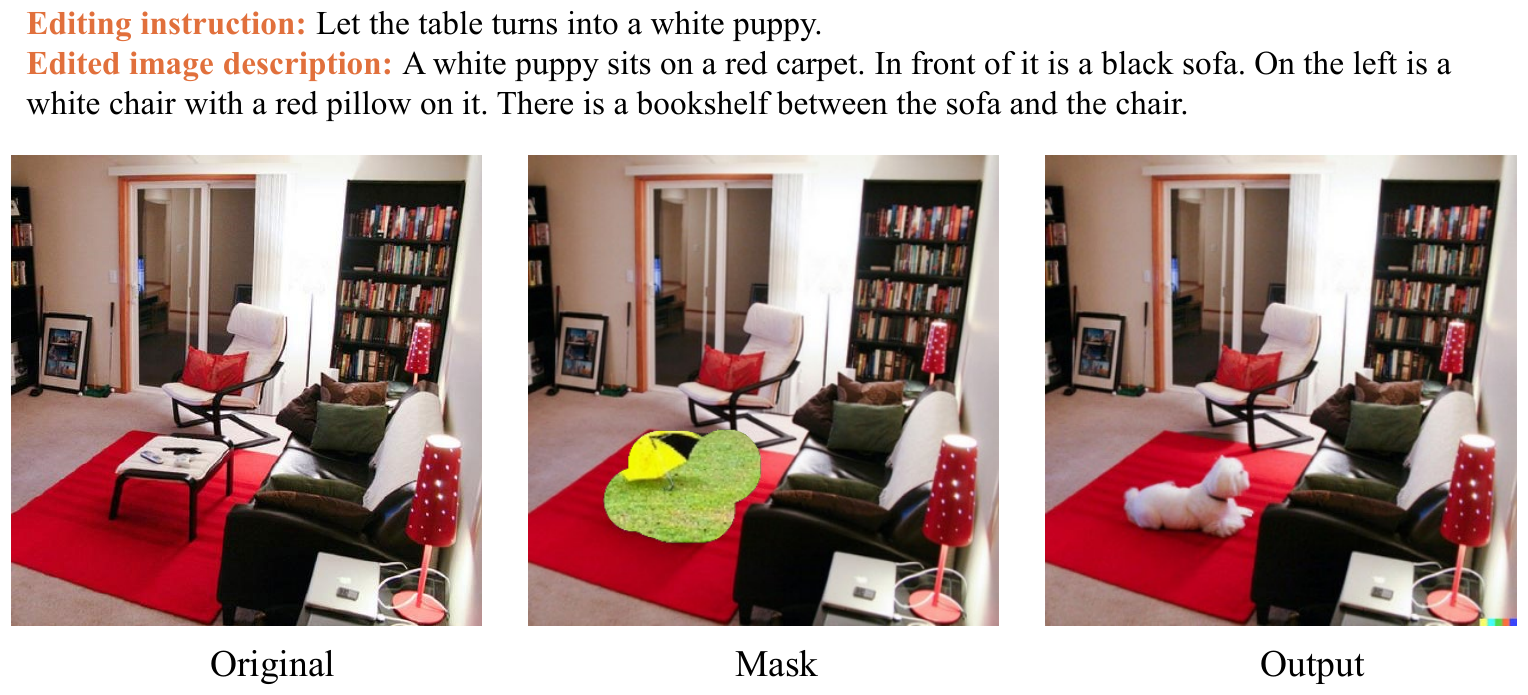}
	\vspace{-4mm}
	\caption{Object Replacement Example I.}
	\label{fig:book2}  
\end{figure}

\begin{figure}[h!]
	\centering  
	\includegraphics[width=1.0\textwidth]{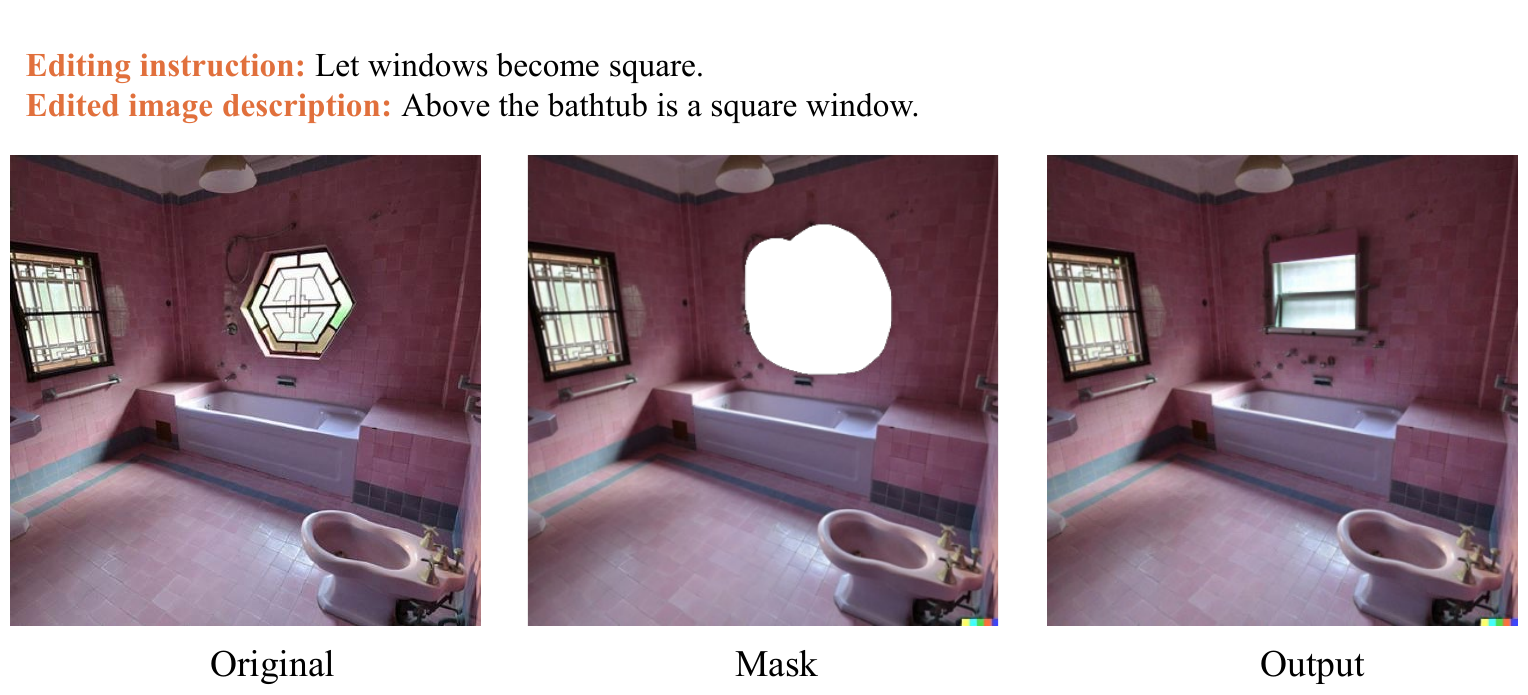}
	\vspace{-4mm}
	\caption{Object Replacement Example II.}
	\label{fig:book3}  
\end{figure}

\textbf{(1.2) Object Replacement}. As shown in Figure~\ref{fig:book2} and Figure~\ref{fig:book3}, we modify the type of an object, change a part of an object, or alter its shape.
\clearpage 

\begin{figure}[h!]
	\centering  
	\vspace{-4mm}
	\includegraphics[width=1.0\textwidth]{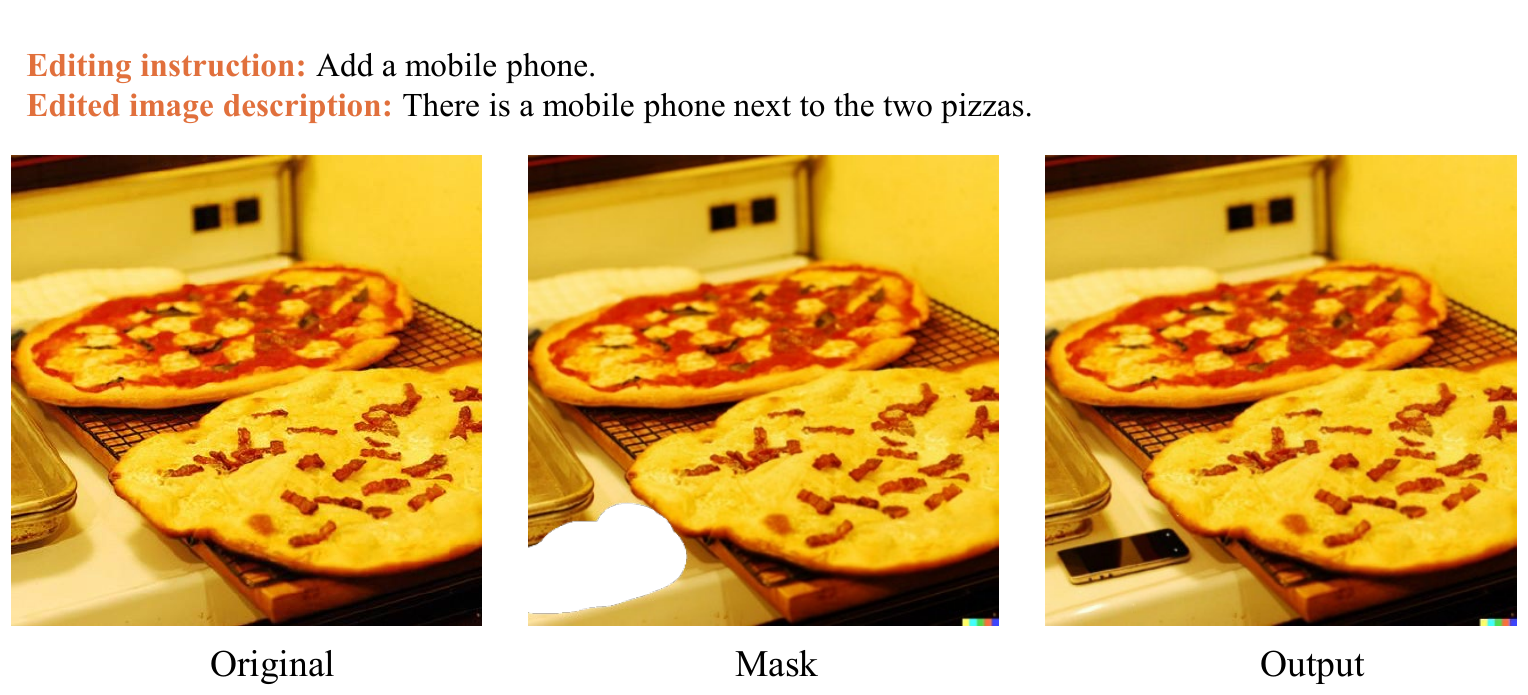}
	\vspace{-4mm}
	\caption{Case of Object Addition.}
	\label{fig:book4}  
\end{figure}

\begin{figure}[h!]
	\centering  
	\vspace{-8mm}
	\includegraphics[width=.94\textwidth]{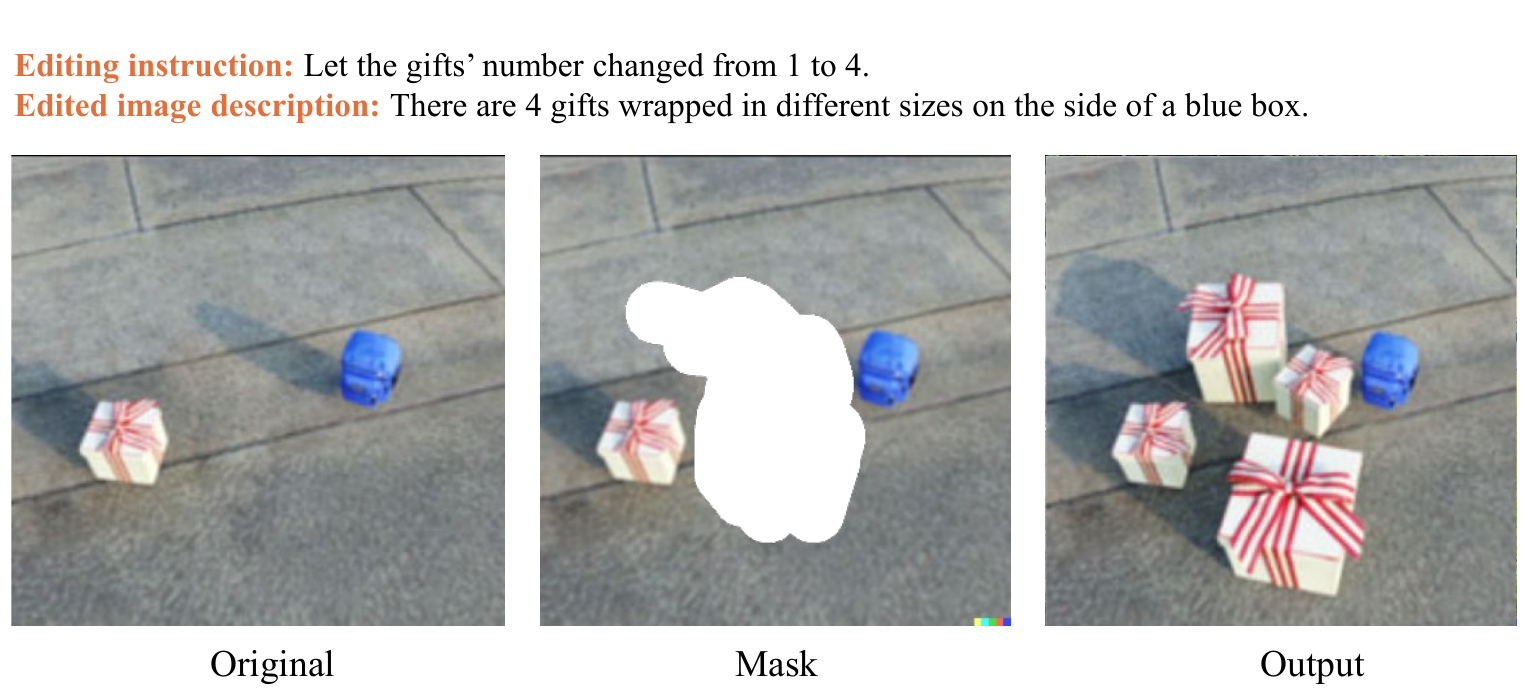}
	\vspace{-4mm}
	\caption{Case of Object Counting Change.}
	\label{fig:book5}  
\end{figure}

\textbf{(1.3) Object Addition}. 
As shown in Figure~\ref{fig:book4}, we add an object to the original image.
\clearpage 
\textbf{(1.4) Object Counting Change}. 
As shown in Figure~\ref{fig:book5}, we can also alter the number of objects in the image. However, it is important to note that the number of objects cannot be reduced to zero (which would be equivalent to removal), nor can it be increased from none to any (which would be considered addition).

\begin{figure}[h!]
	\centering  
	\vspace{-4mm}
	\includegraphics[width=.94\textwidth]{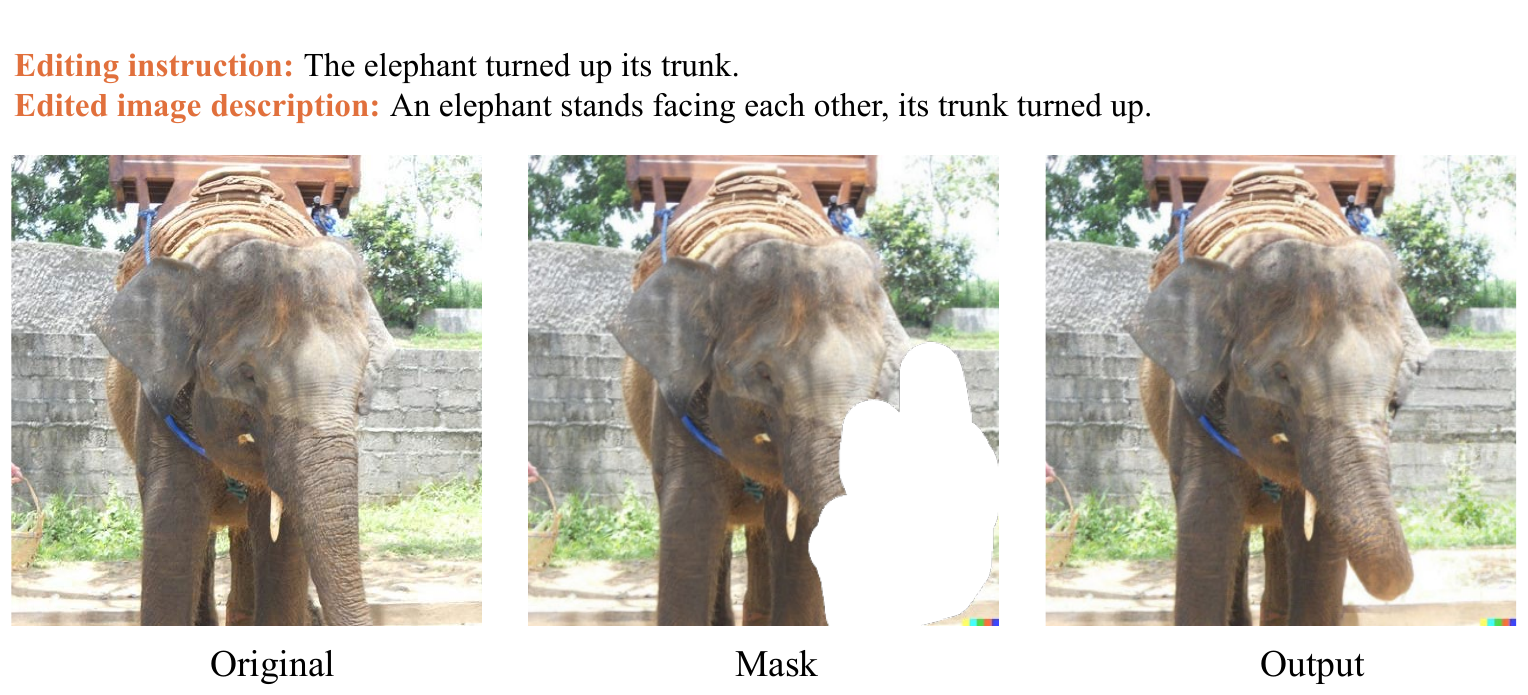}
	\vspace{-4mm}
	\caption{Case of Action Change.}
	\label{fig:book6}  
\end{figure}

\textbf{(2) Action Change}. 
As shown in Figure~\ref{fig:book6}, if the subject is a specific organism, its actions can also be altered.

\begin{figure}[h!]
	\centering  
	\vspace{-4mm}
	\includegraphics[width=.94\textwidth]{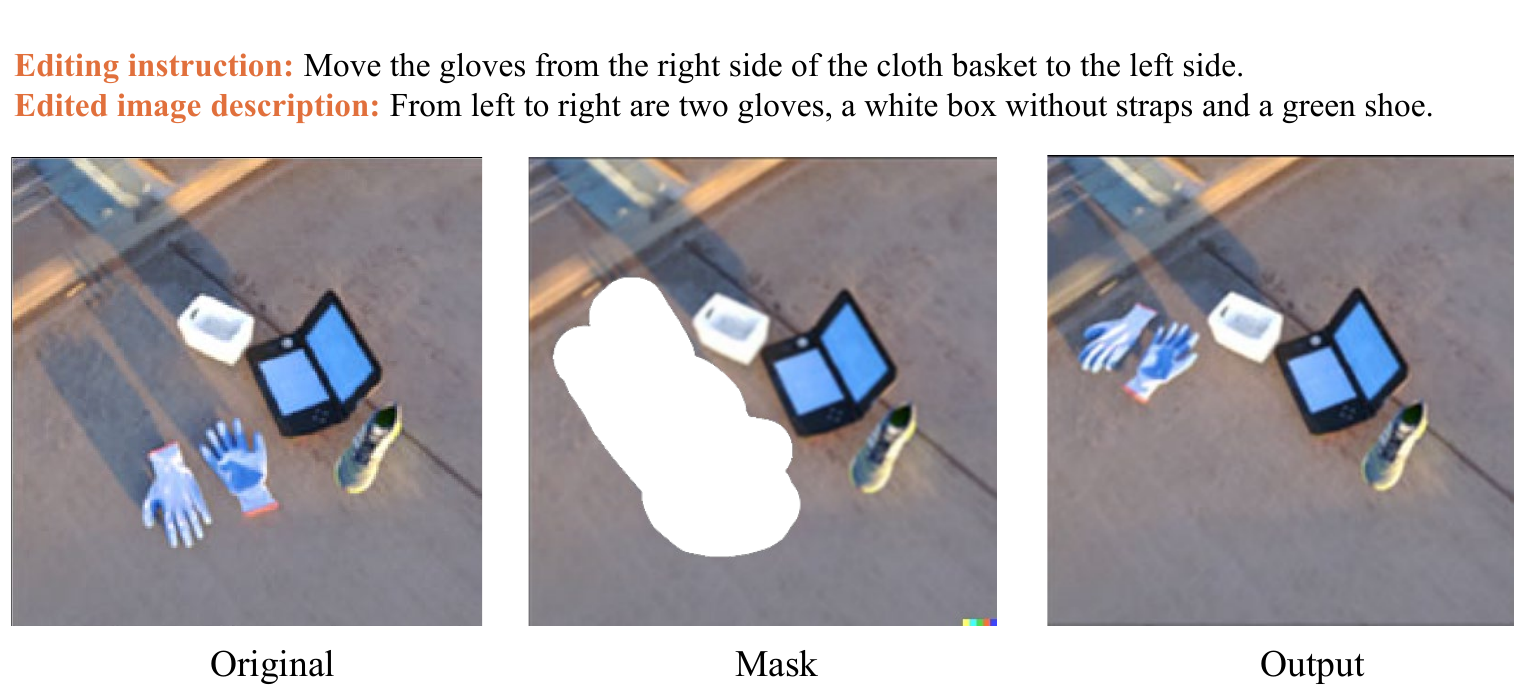}
	\vspace{-4mm}
	\caption{Case of Relation Change.}
	\label{fig:book7}  
\end{figure}

\textbf{(3) Relation Change}. 
As shown in Figure~\ref{fig:book7}, another type of editing involves modifying the relationships between objects.
\clearpage 

\subsection{Notes for Annotators}
\textbf{(1) Selection of Prompt Words}. When using DALL-E 2, if only an editing instruction (as shown in Figure 1) is provided, the model’s generated results are often poor. It is recommended to use a detailed description of the target image (as shown in Figure 2). For example:  
Editing instruction: \textit{"Let the boy turn into a girl."}  

\begin{figure}[h!]
	\centering  
	\includegraphics[width=1.0\textwidth]{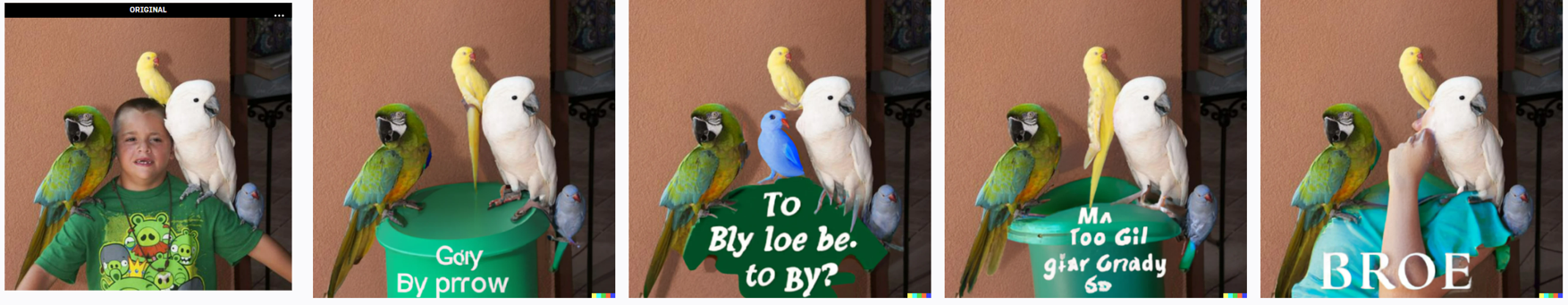}
	\vspace{-4mm}
	\caption{An Example of Prompt Word Selection}
	\label{fig:note1}  
\end{figure}

Target Image Caption: \textit{Four parrots are perched on a girl's shoulders, arms, and head.}
\begin{figure}[h!]
	\centering  
	\includegraphics[width=1.0\textwidth]{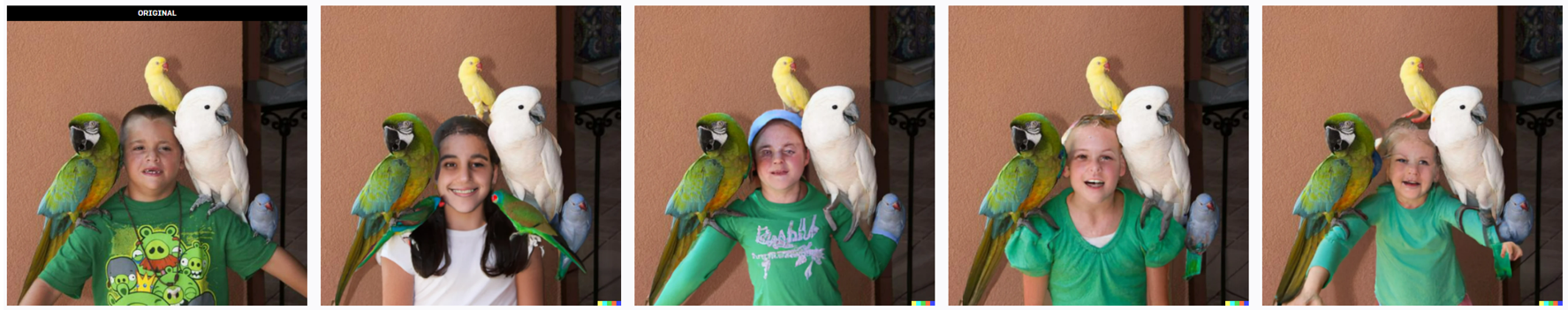}
	\vspace{-4mm}
	\caption{An Example of Prompt Word Selection}
	\label{fig:note2}  
\end{figure}

\textbf{(2) Image Resolution}.
After uploading the image, click 'crop' first, then click 'Edit image' to proceed with editing.

\begin{figure}[h!]
	\centering  
	\includegraphics[width=0.94\textwidth]{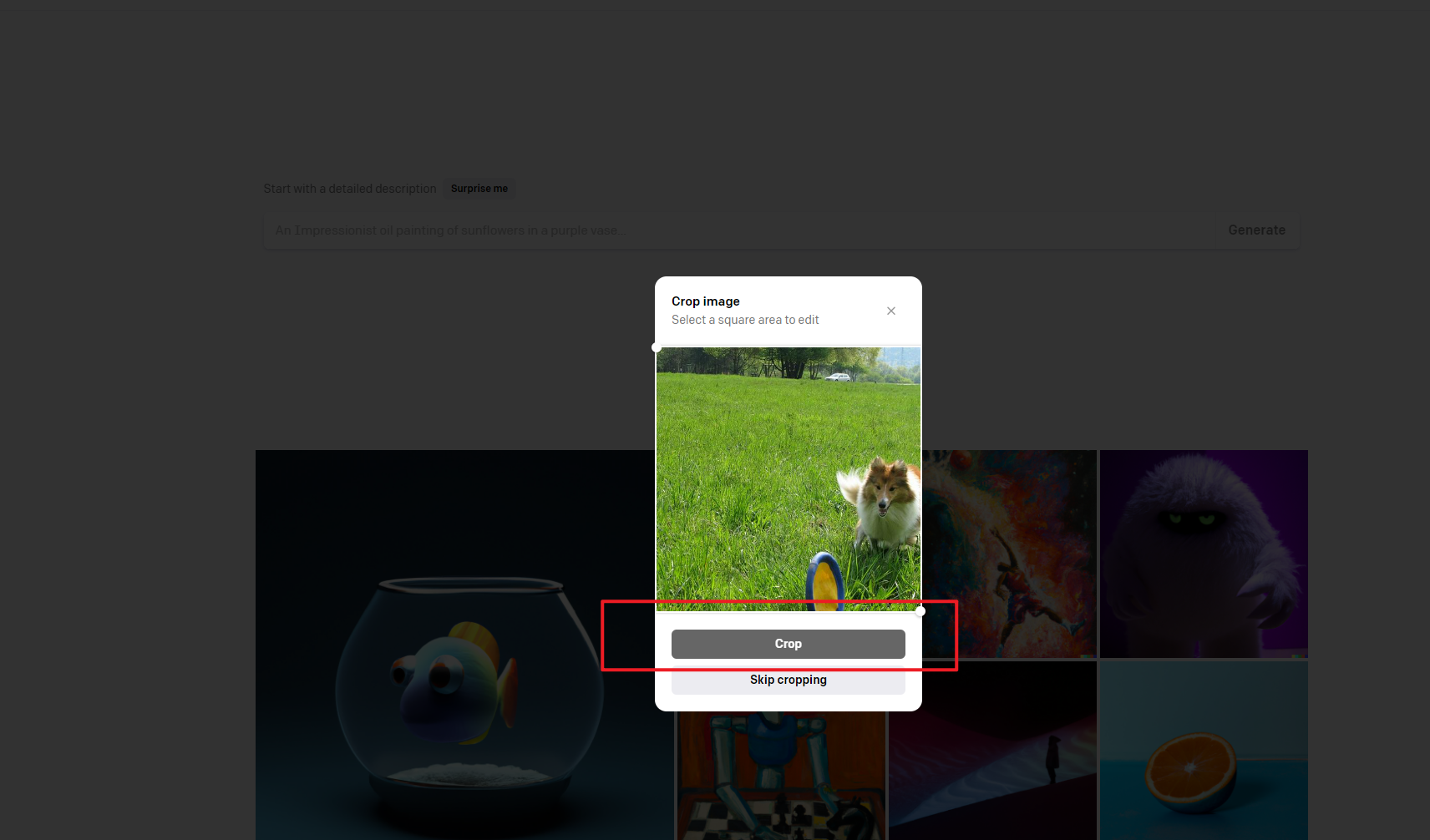}
	\caption{Performing a Crop Operation on the DALL-E 2 Platform.}
	\label{fig:note3}  
\end{figure}
\clearpage 
\begin{figure}[h!]
	\centering  
	\vspace{4mm}
	\includegraphics[width=0.94\textwidth]{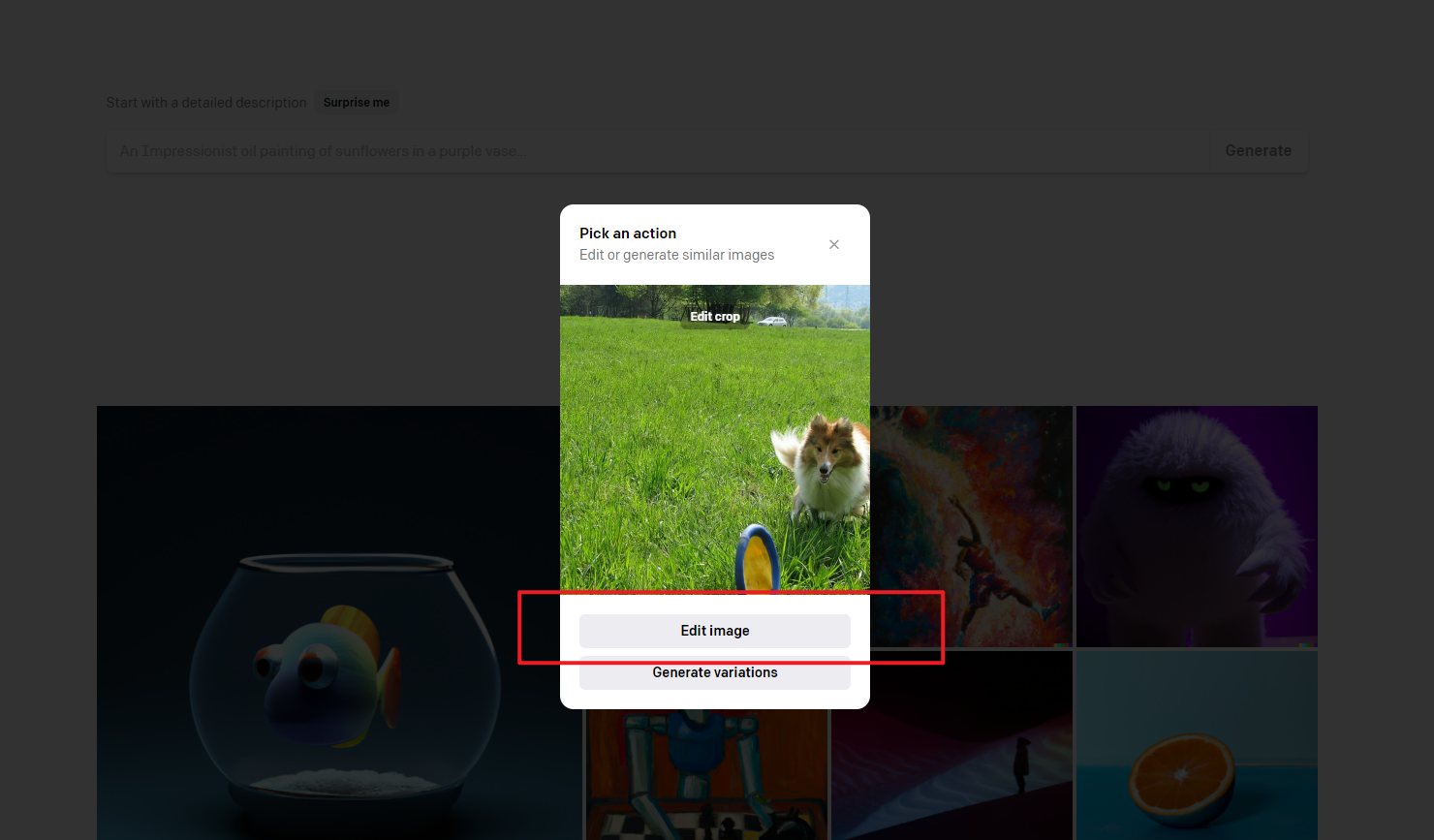}
	\caption{Performing an Editing Operation on the DALL-E 2 Platform.}
	\label{fig:note4}  
\end{figure}

\textbf{(3) Avoid Editing Irrelevant Areas}.
When masking areas, avoid using too large a mask, as this can lead to distortion or result in editing that does not cover the intended area. For example, in the Figure~\ref{fig:note5} below, the boat paddle disappears, which is unreasonable.  
\begin{figure}[h!]
	\centering  
	\includegraphics[width=1.0\textwidth]{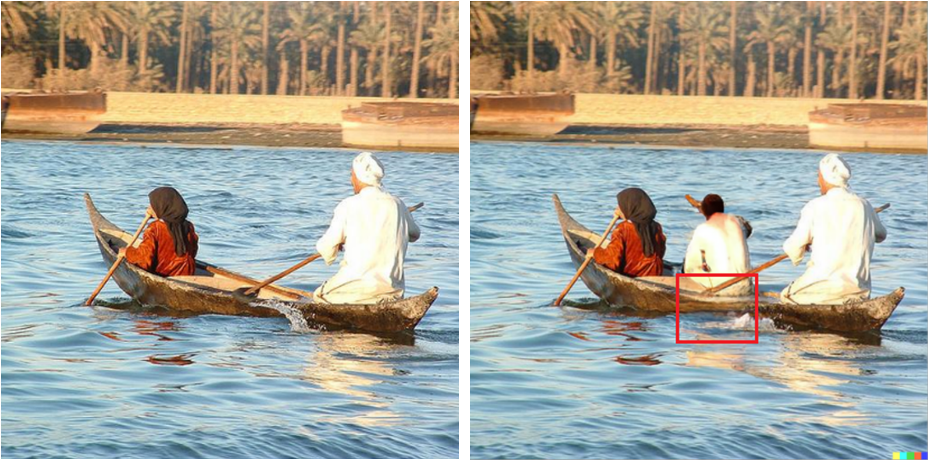}
	\vspace{-4mm}
	\caption{An Illustration of Avoiding Edits in Irrelevant Areas.}
	\label{fig:note5}  
\end{figure}

\begin{figure}[h!]
	\centering  
	\vspace{14mm}
	\includegraphics[width=1.0\textwidth]{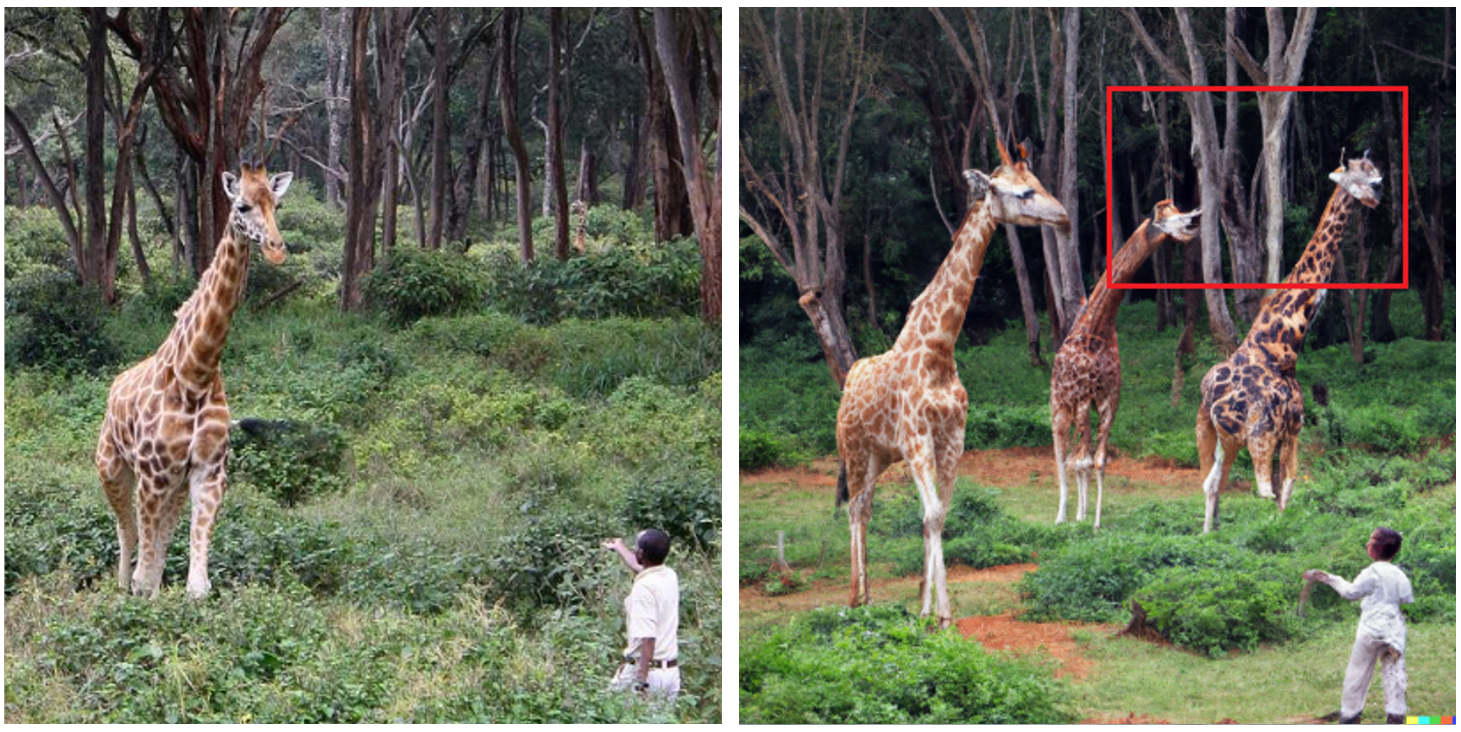}
	\vspace{-4mm}
	\caption{An Illustration of Avoiding Edits in Irrelevant Areas.}
	\label{fig:note6}  
\end{figure}

Additionally, if the editing task is to add a giraffe, the expected result should be the addition of two giraffes as shown in Figure~\ref{fig:note6}. However, the output image shows excessive changes (likely due to an overly large mask area). A reminder: the mask area should not be too large; it should be appropriate.  
Also, the giraffe's head in this example is generated unrealistically.  
\clearpage 

\begin{figure}[h!]
	\centering  
	\includegraphics[width=1.0\textwidth]{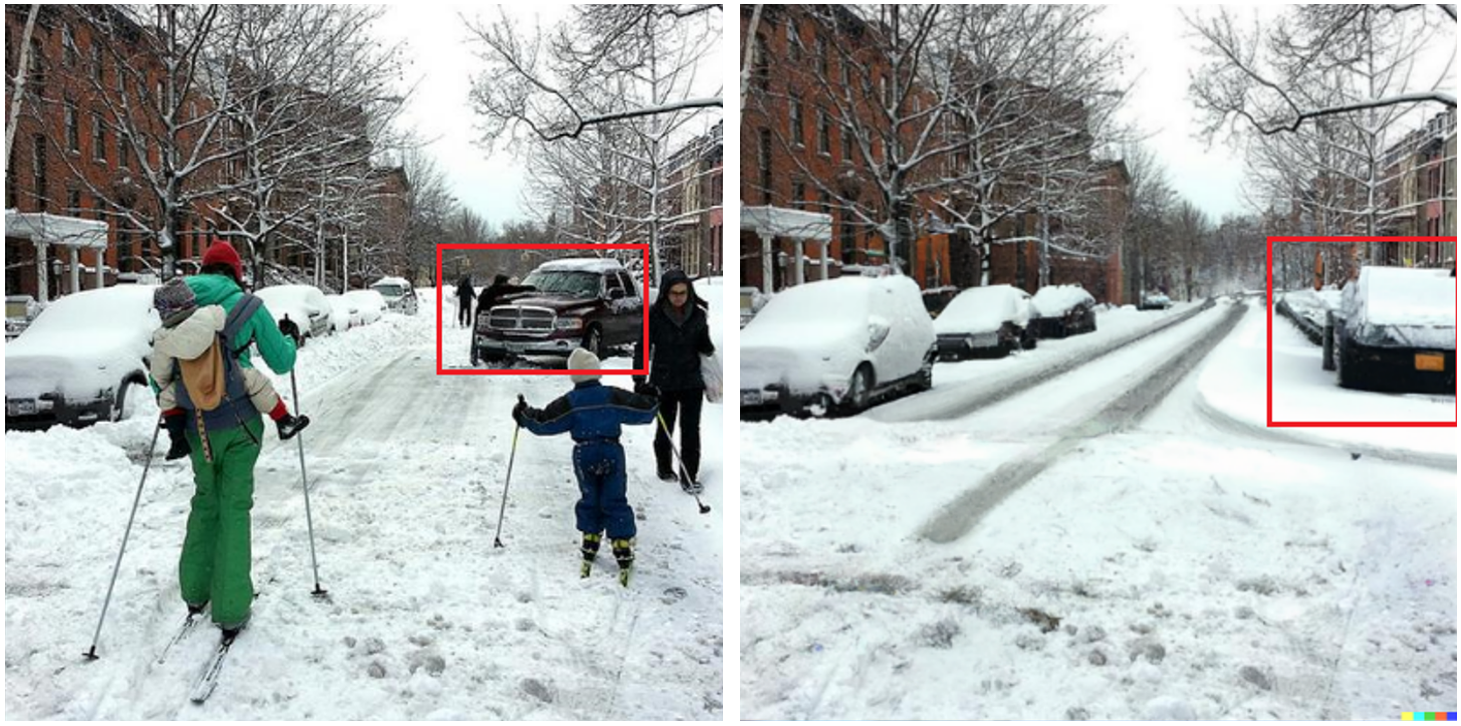}
	\vspace{-4mm}
	\caption{An Illustration of Avoiding Edits in Irrelevant Areas.}
	\label{fig:note7}  
\end{figure}

When the instruction is to remove a person, it is best not to change the car for Figure~\ref{fig:note7}.

\begin{figure}[h!]
	\centering  
	\includegraphics[width=1.0\textwidth]{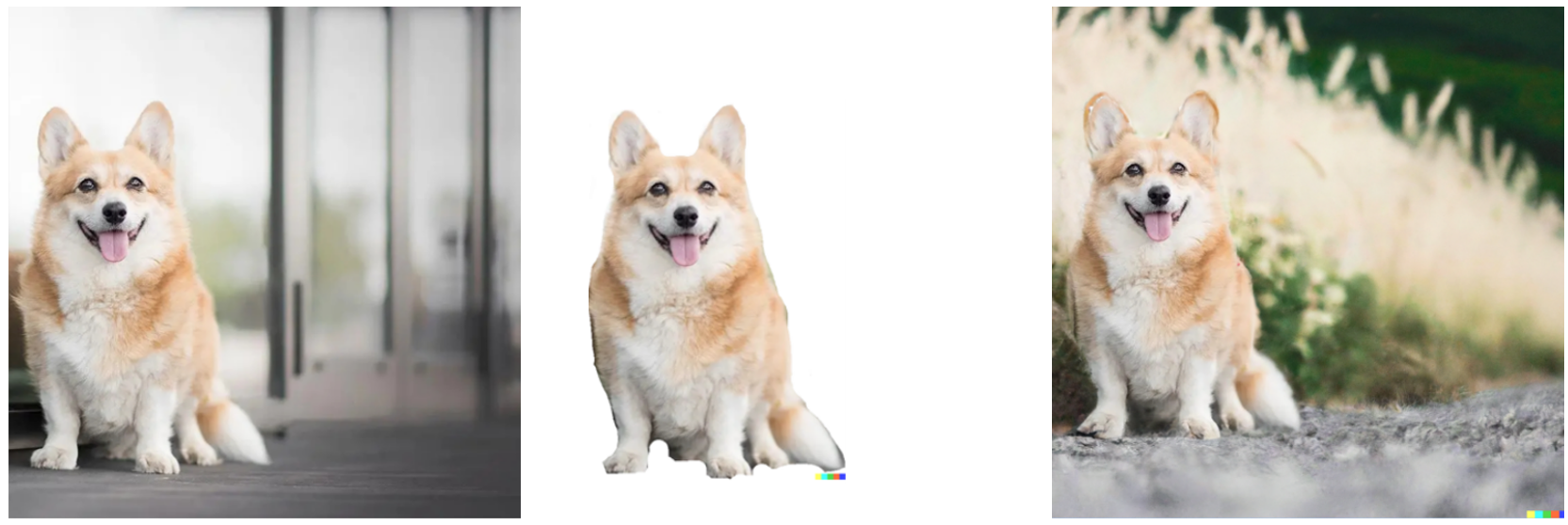}
	\vspace{-4mm}
	\caption{An Illustration of Avoiding Edits in Irrelevant Areas.}
	\label{fig:note8}  
\end{figure}

The following masking as shown in Figure~\ref{fig:note8} is done well: the instruction is to change the background, and everything except for the dog is masked.

\clearpage 

\begin{figure}[h!]
	\centering  
	\vspace{14mm}
	\includegraphics[width=1.0\textwidth]{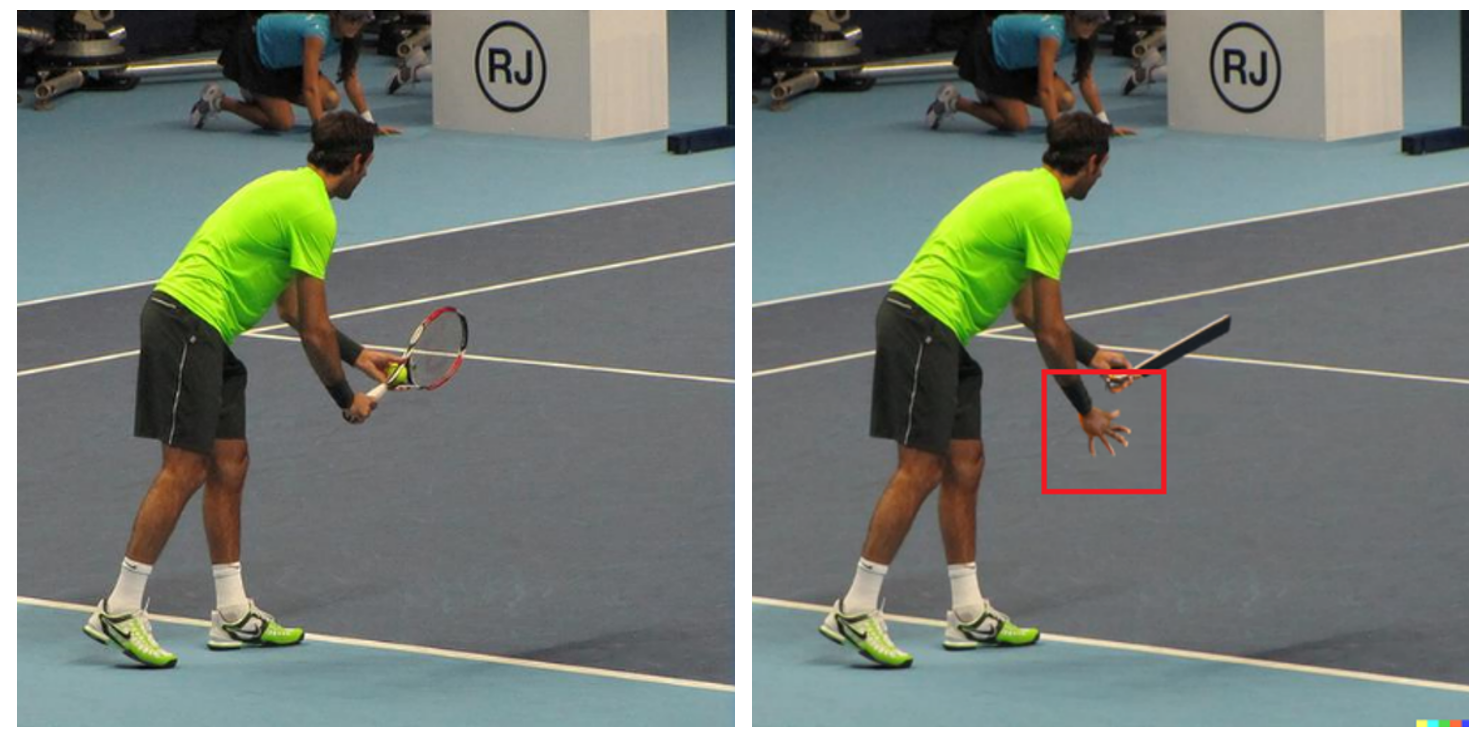}
	\vspace{-4mm}
	\caption{A Case for Ensuring Edit Quality.}
	\label{fig:note9}  
\end{figure}

\textbf{(4) Quality of Edits Ensure}.
DALL-E 2 sometimes struggles to interpret instructions accurately, so attention to detail in editing structures is important. For example, in the following case shown in Figure~\ref{fig:note9}, the fingers are distorted and do not resemble a normally outstretched hand.

\begin{figure}[h!]
	\centering  
	\vspace{6mm}
	\includegraphics[width=1.0\textwidth]{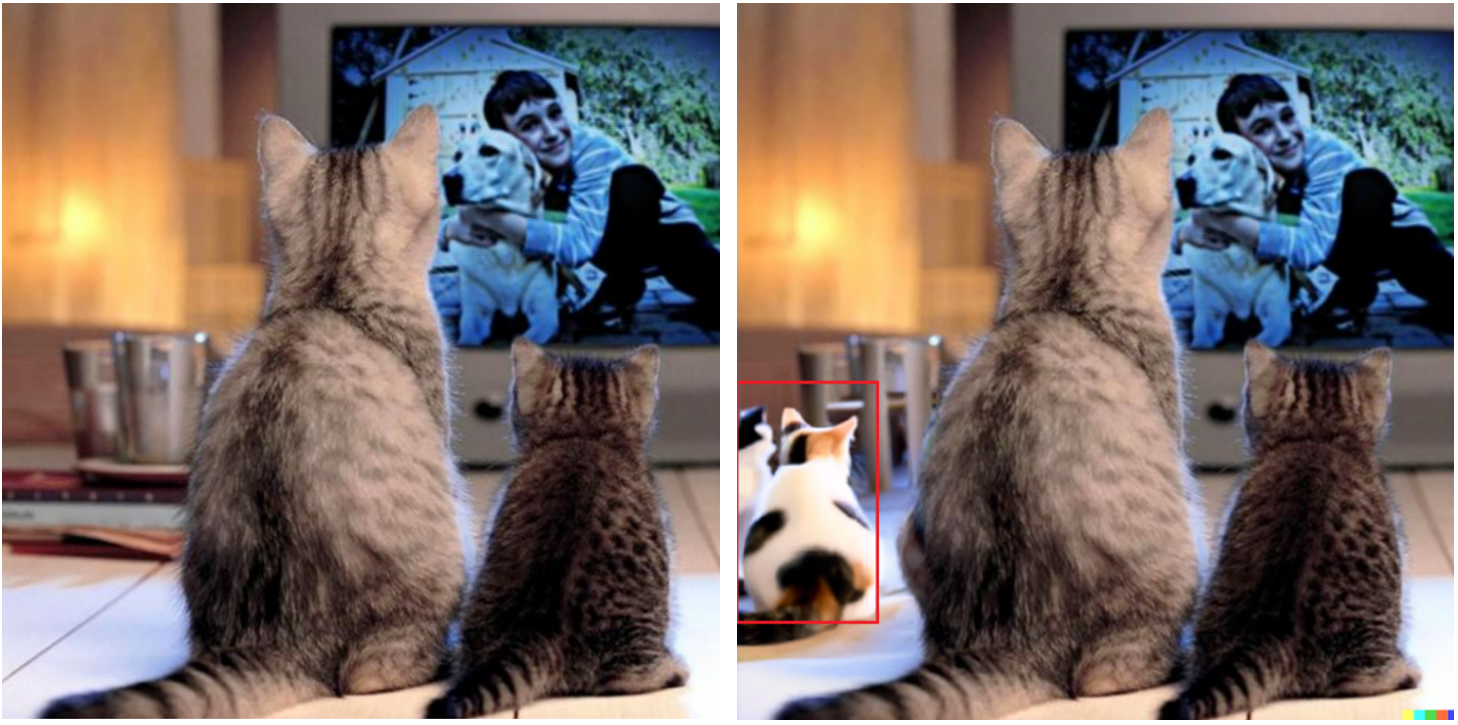}
	\vspace{-4mm}
	\caption{A Case for Ensuring Edit Quality.}
	\label{fig:note10}  
\end{figure}

As demonstrated in Figure~\ref{fig:note10} , the image description is \textit{"The back view of a large calico cat sitting next to two other cats,"} but the actual image shows four cats.
\clearpage 

\begin{figure}[h!]
	\centering  
	\includegraphics[width=1.0\textwidth]{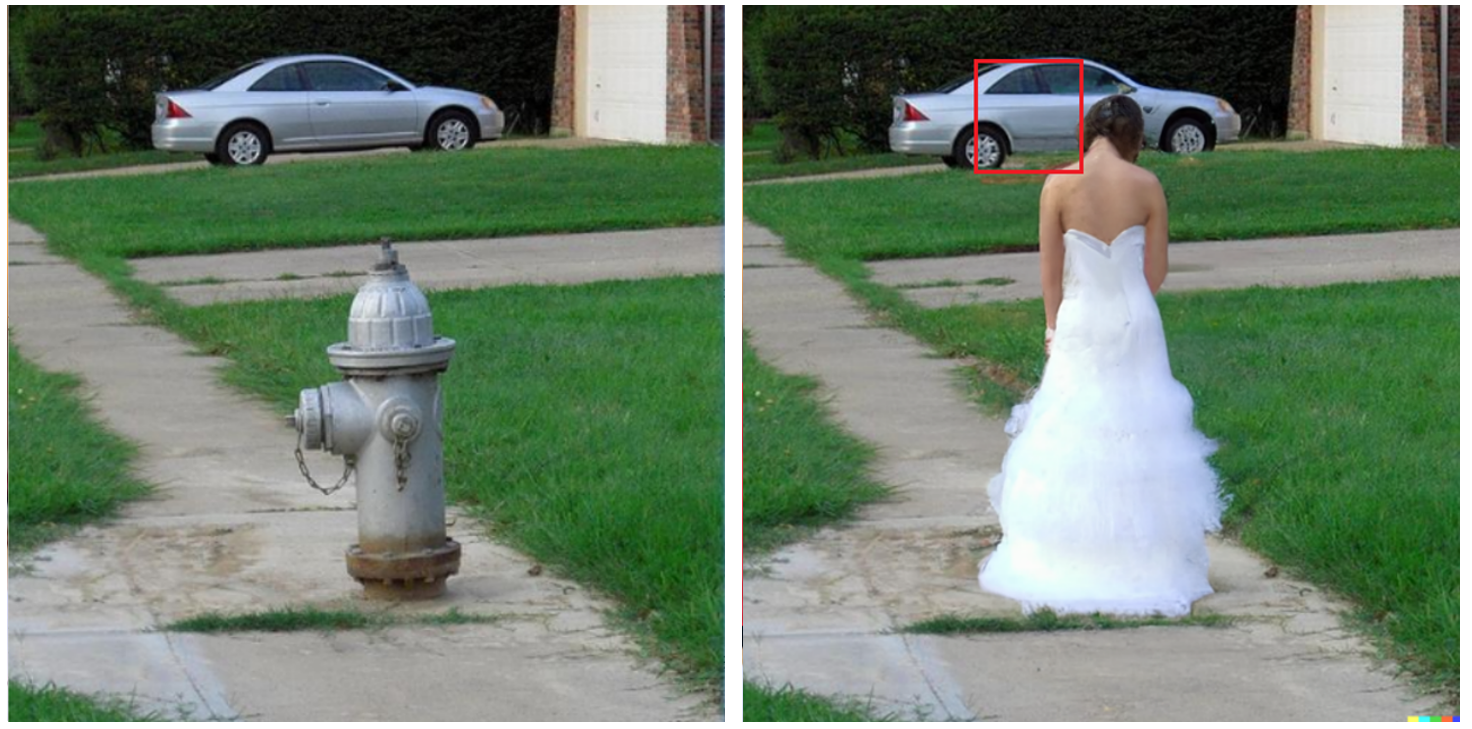}
	\vspace{-4mm}
	\caption{A Case for Ensuring Edit Quality.}
	\label{fig:note11}  
\end{figure}

The car door in Figure~\ref{fig:note11} has disappeared, which is also unreasonable (this issue was caused by an overly large masked area).

\textbf{(5) Success Rate}.
DALL-E 2 has a relatively low success rate. If multiple regenerations or instruction modifications do not yield satisfactory results, it may be best to abandon the task. The exact number of attempts before abandonment is left to the discretion of the annotator. For simplicity, only one editing instruction should be tried for each image, and the best result should be selected.

\textbf{(6) Consistency in Style Before and After Editing}.
If the original image is black and white, the edited result should also be in black and white.  
Generally, DALL-E 2’s generated results tend to adhere to the original style, so there is no need to explicitly guide the editing in terms of style. However, attention should be paid when selecting the final result to ensure consistency.
\clearpage 
\begin{figure}[h!]
	\centering  
	\vspace{-4mm}
	\includegraphics[width=1.0\textwidth]{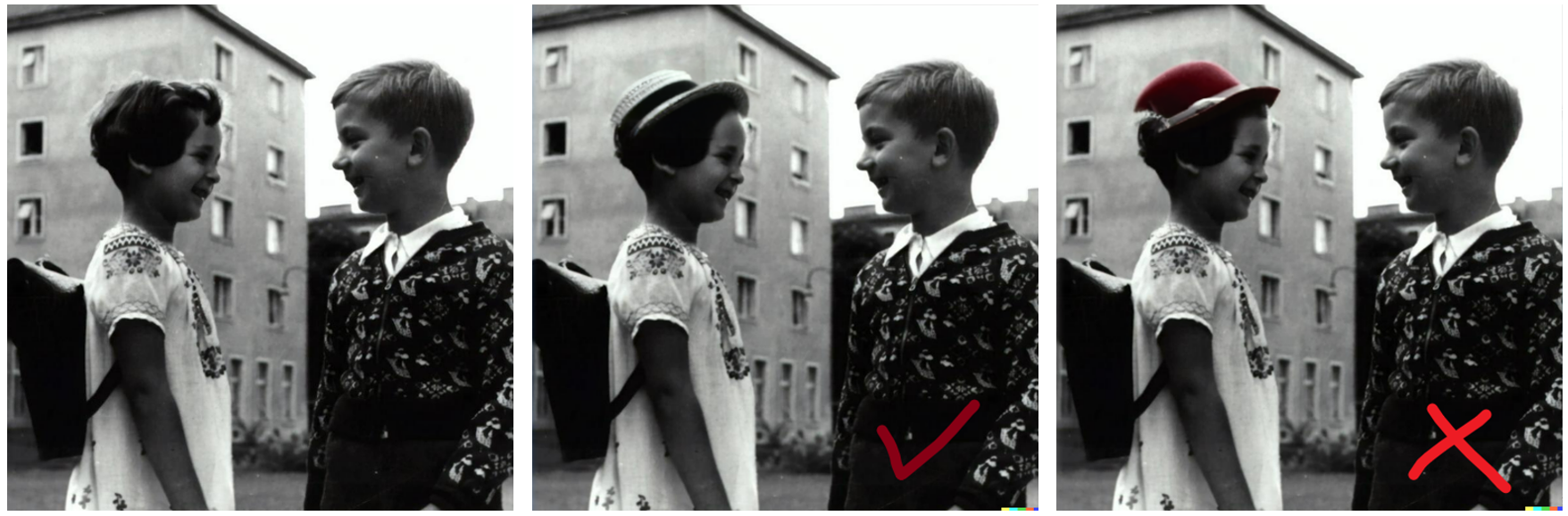}
	\vspace{-4mm}
	\caption{An Illustration of Consistency in Style Before and After Editing.}
	\label{fig:note12}  
\end{figure}

\subsection{Initial Image Selection}
As mentioned in Section~\ref{sec:pipe}, we implement a rigorous selection process to ensure the quality of the original images. It is important to note that, at the beginning of the annotation process, annotators are still given the opportunity to reselect the original image. Figure~\ref{fig:note13} below illustrates an example of the selection process. For instance, image (a) is acceptable, while (b) contains some unusual artifacts, (c) has poor image quality, and (d) has low resolution and lacks sufficient visual information.

\begin{figure}[h!]
	\centering  
	\vspace{-2mm}
	\includegraphics[width=1.0\textwidth]{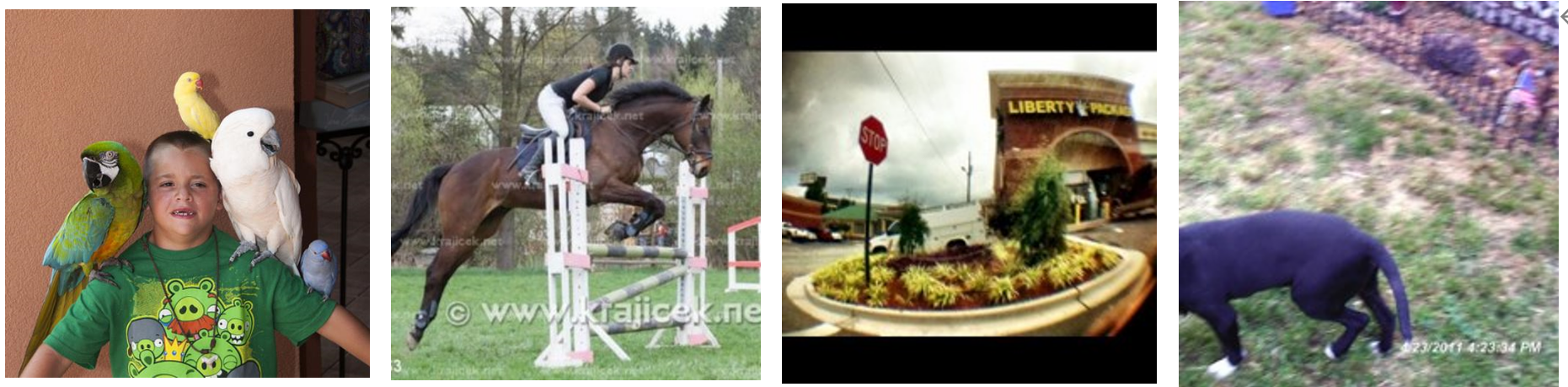}
	\vspace{-4mm}
        \caption{Examples of valid and invalid images. The first image is valid, while the following three images are invalid.}
	\label{fig:note13}  
\end{figure}

\subsection{Image Editing Process and Annotation Platform}
\textbf{(1) Log in to the DALL·E 2 platform and click "Try DALL-E" to upload an image.}
\begin{figure}[h!]
	\centering  
	\vspace{-1mm}
	\includegraphics[width=1.0\textwidth]{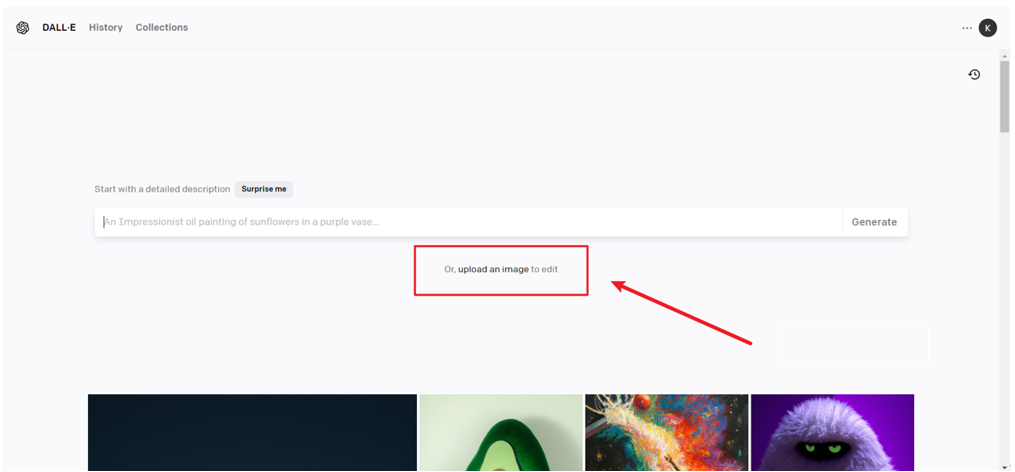}
	\vspace{-4mm}
        \caption{Log in to the DALL·E 2 platform and click "Try DALL-E" to upload an image.}
	\label{fig:pipe1}  
\end{figure}
\clearpage 
\vspace{8mm}
\textbf{(2) After uploading the image, a cropping page will be displayed.}
\begin{figure}[h!]
	\centering  
	\includegraphics[width=1.0\textwidth]{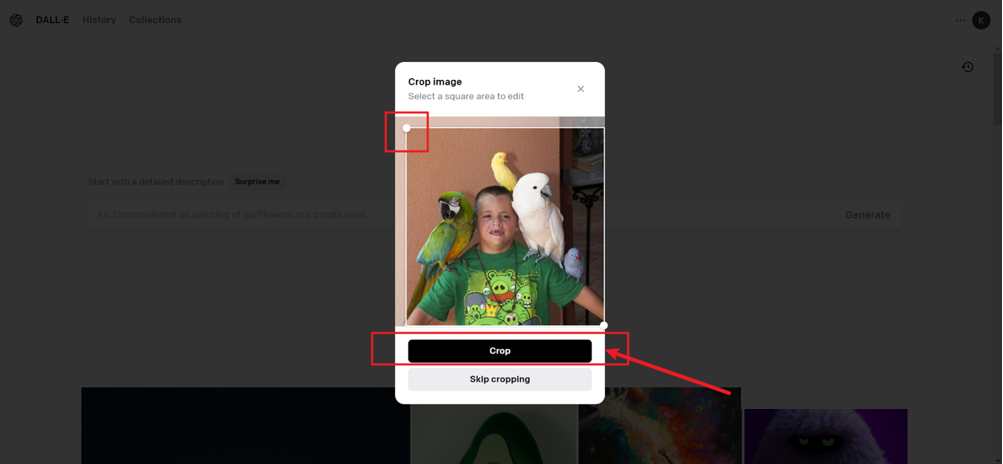}
        \caption{After uploading the image, a cropping page will be displayed.}
	\label{fig:pipe2}  
\end{figure}

\textbf{(3) Click the "Edit" button to enter the editing window.}
\begin{figure}[h!]
	\centering  
	\includegraphics[width=1.0\textwidth]{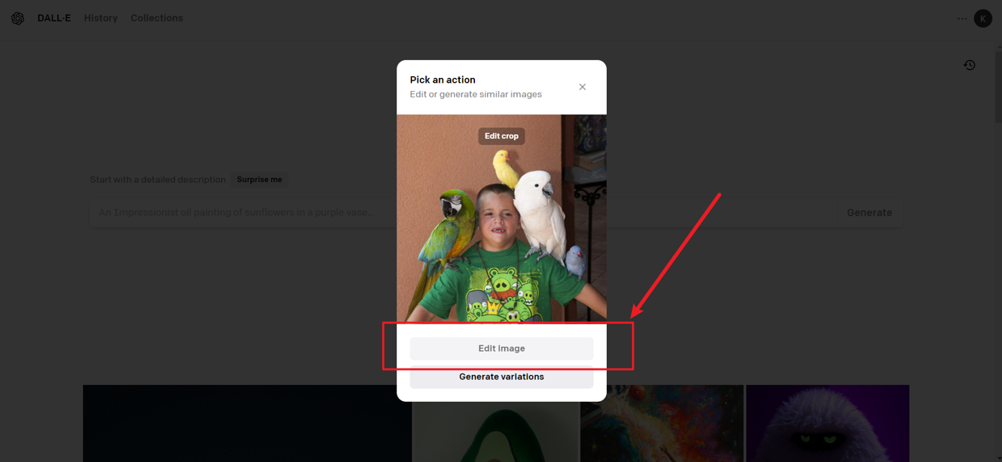}
        \caption{Click the "Edit" button to enter the editing window.}
	\label{fig:pipe3}  
\end{figure}

\clearpage 
\vspace{8mm}
\textbf{(4) Drag the editing points to select the area to be edited.}
\begin{figure}[h!]
	\centering  
	\includegraphics[width=1.\textwidth]{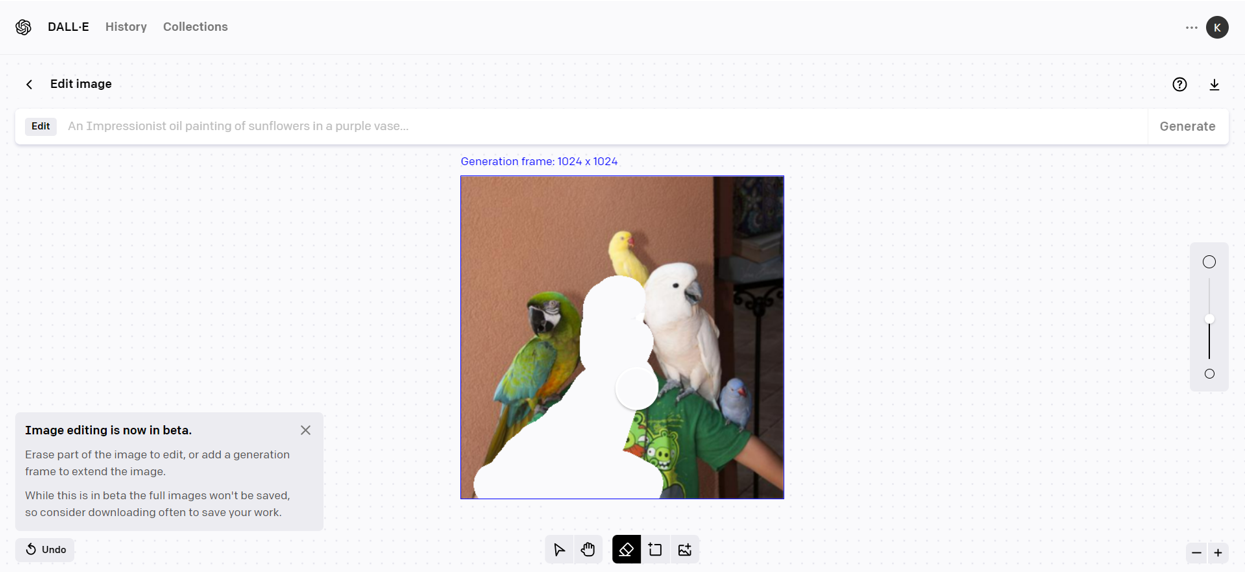}
	\vspace{-3mm}
        \caption{Drag the editing points to select the area to be edited.}
	\label{fig:pipe4}  
\end{figure}

Next, input the editing instructions in the text box. For example, if your task is to change an object, first select the person, and then define the editing instruction as “change the boy into a girl.” At this point, combine the previously selected description and imagine the expected edited image (the more detailed the description, the better), and enter it in the text box. For example, “Four parrots are perched on a cute girl’s arms and shoulders” (Note: the output box for editing instructions will only appear after selecting the editing contours). Then save the mask and choose to have the model generate the image.

\begin{figure}[h!]
	\centering  
	\includegraphics[width=1.\textwidth]{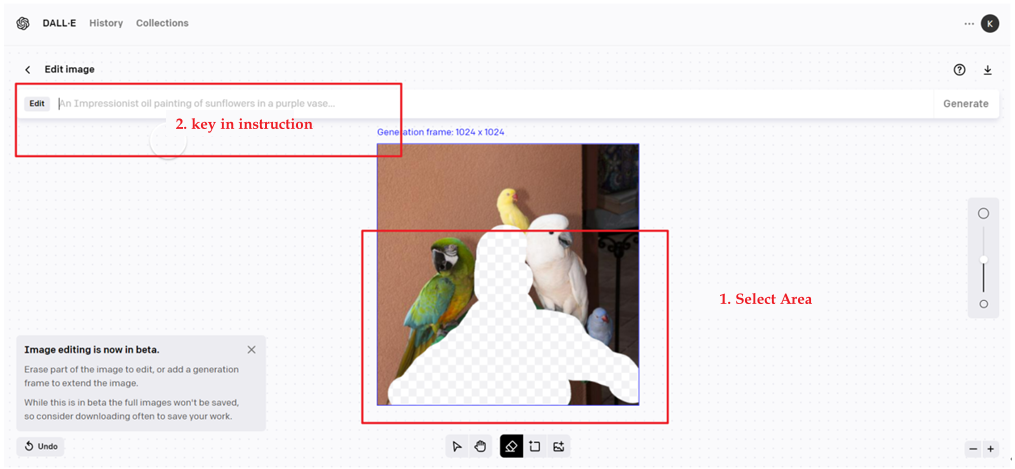}
	\vspace{-3mm}
        \caption{Input the editing instructions in the text bo.}
	\label{fig:pipe5}  
\end{figure}
\clearpage 
\begin{figure}[h!]
	\centering  
	\vspace{-8mm}
	\includegraphics[width=1.\textwidth]{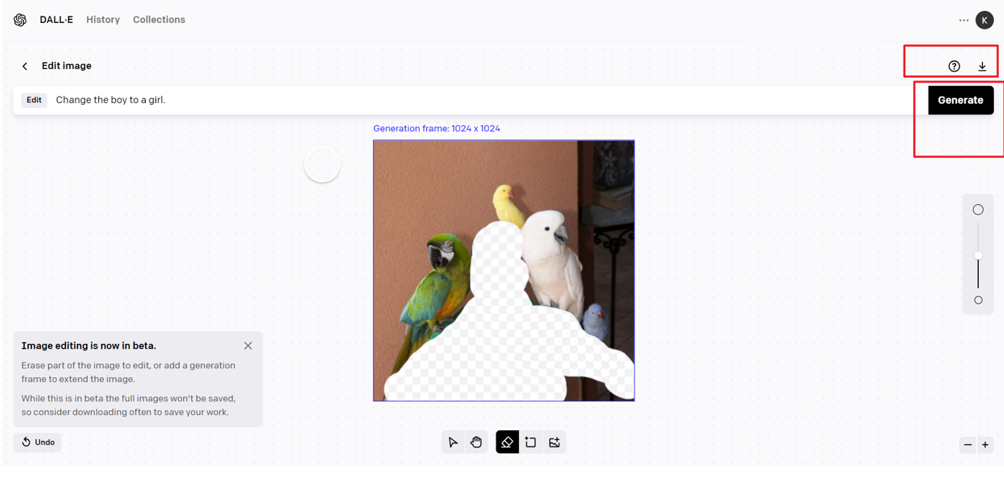}
	\vspace{-3mm}
        \caption{Generate edited images.}
	\label{fig:pipe6}  
\end{figure}

If the generated result is of poor quality (e.g., none of the images meet the requirements), you can click the “regenerate” button to try again.
\begin{figure}[h!]
	\centering  
	\vspace{-3mm}
	\includegraphics[width=1.\textwidth]{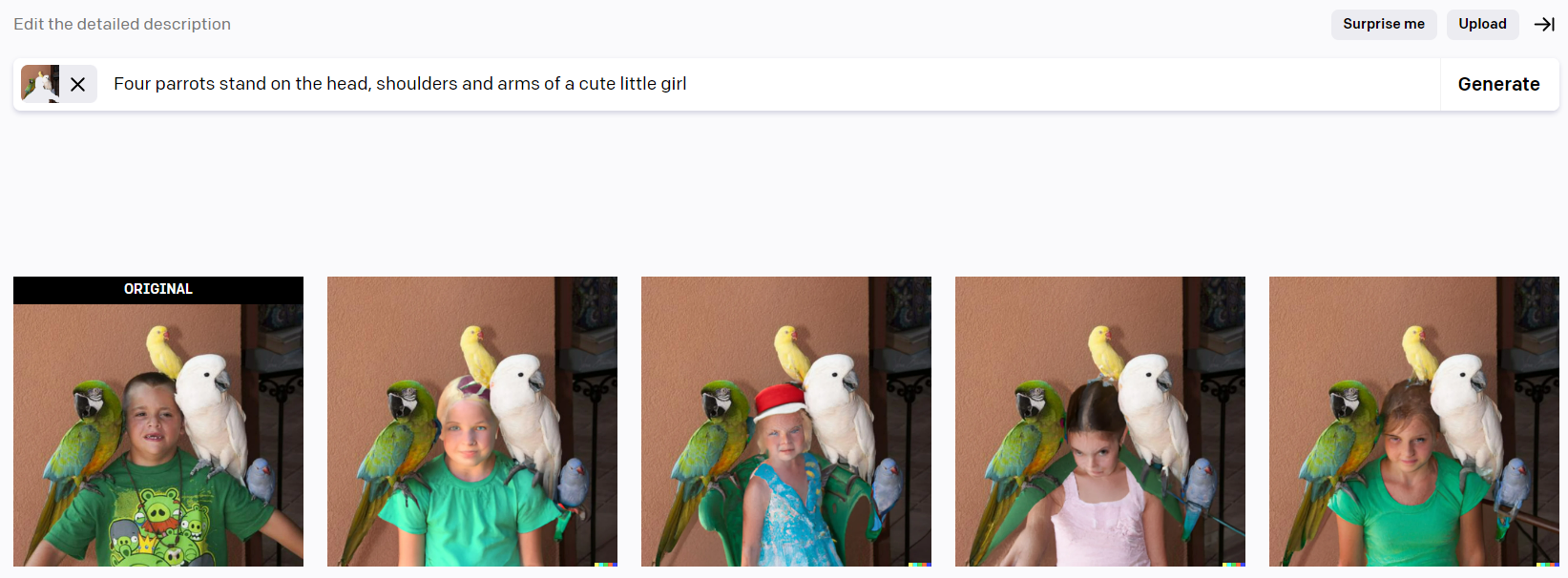}
	\vspace{-3mm}
        \caption{Regenerate edited images.}
	\label{fig:pipe7}  
\end{figure}

However, please avoid generating the same instruction more than three times. Instead, try modifying the instruction to make it more precise. For example, change the expected image description to “A cute little girl with her arms outstretched, with four parrots perched on her head, shoulders, and arms.”

\begin{figure}[h!]
	\centering  
	\vspace{-3mm}
	\includegraphics[width=1.\textwidth]{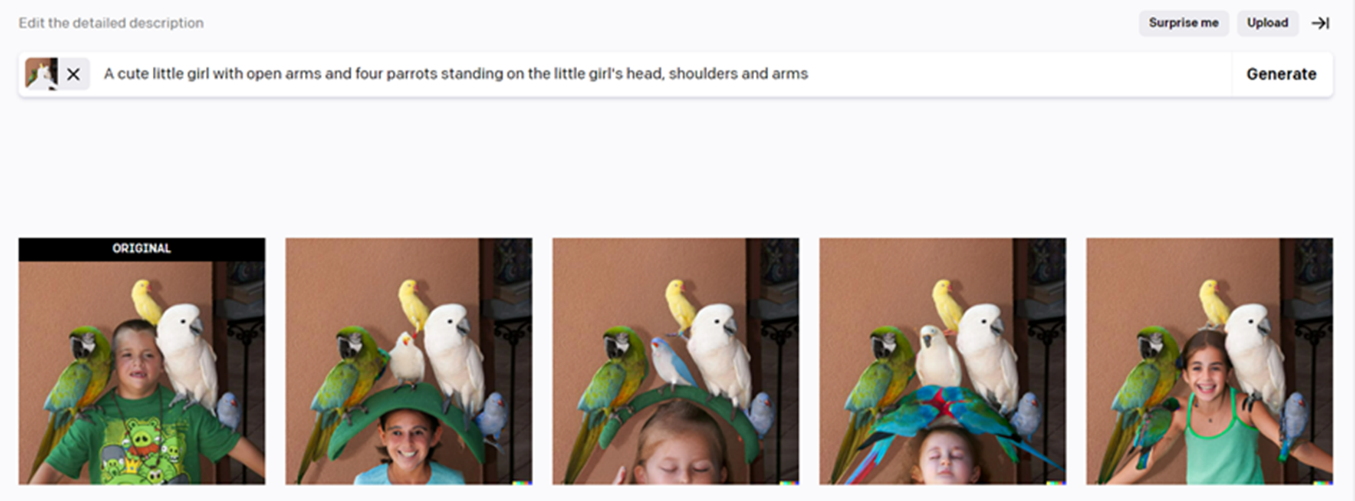}
	\vspace{-3mm}
            \caption{Regenerated images are still not satisfactory and may require revised instructions.}
	\label{fig:pipe8}  
\end{figure}

\clearpage 
Once you find an image that seems appropriate, click on it to download and finish the editing process.
\begin{figure}[h!]
	\centering  
	\vspace{-3mm}
	\includegraphics[width=1.\textwidth]{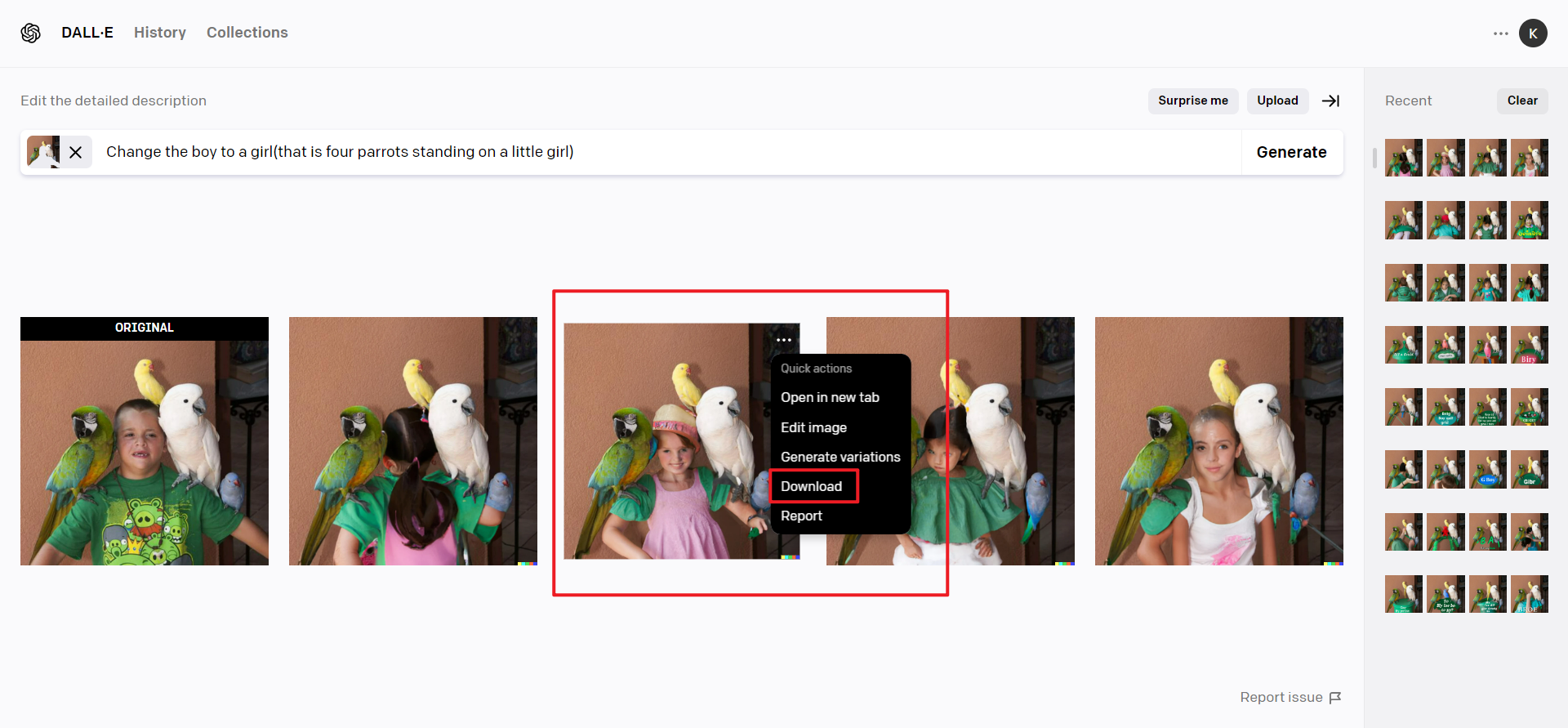}
	\vspace{-3mm}
            \caption{Download and finish the editing process.}
	\label{fig:pipe9}  
\end{figure}

\begin{figure}[h!]
	\centering  
	\vspace{-3mm}
	\includegraphics[width=1.\textwidth]{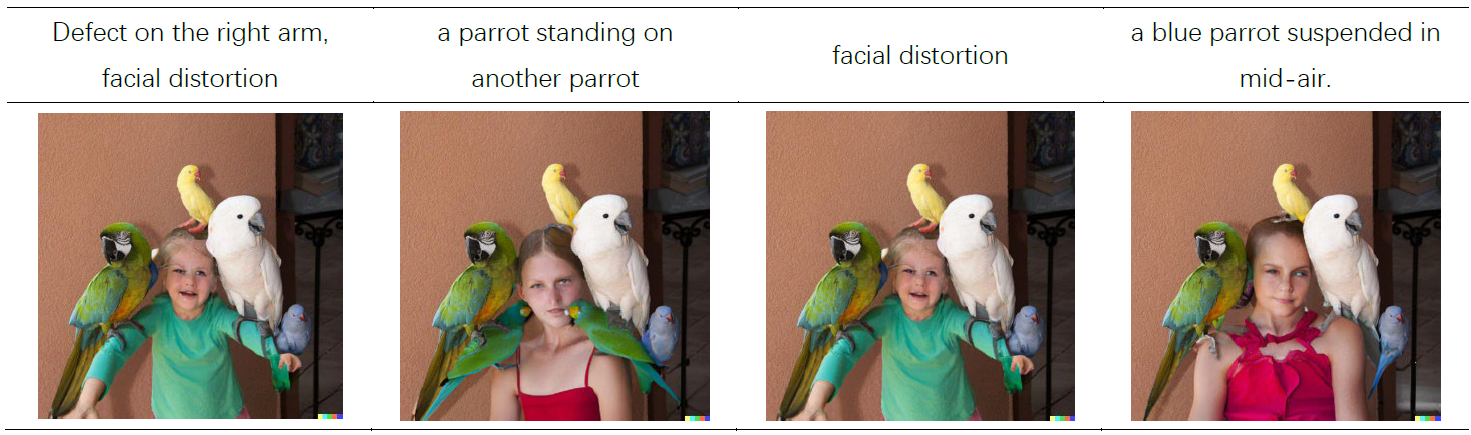}
	\vspace{-3mm}
            \caption{Defective Image Example.}
	\label{fig:pipe10}  
\end{figure}

\textbf{(5) Result selection}. Please ensure that the final selected image is semantically accurate and as realistic as possible, without significant flaws. Below are some examples of poor results in Figure~\ref{fig:pipe10}. Please try to avoid these mistakes, as we will consider instructions that result in such issues as non-compliant.

\textbf{(6) Submission of results}. Finally, you need to submit the following materials as a group to our platform.
\begin{figure}[h!]
	\centering  
	\vspace{-3mm}
	\includegraphics[width=1.\textwidth]{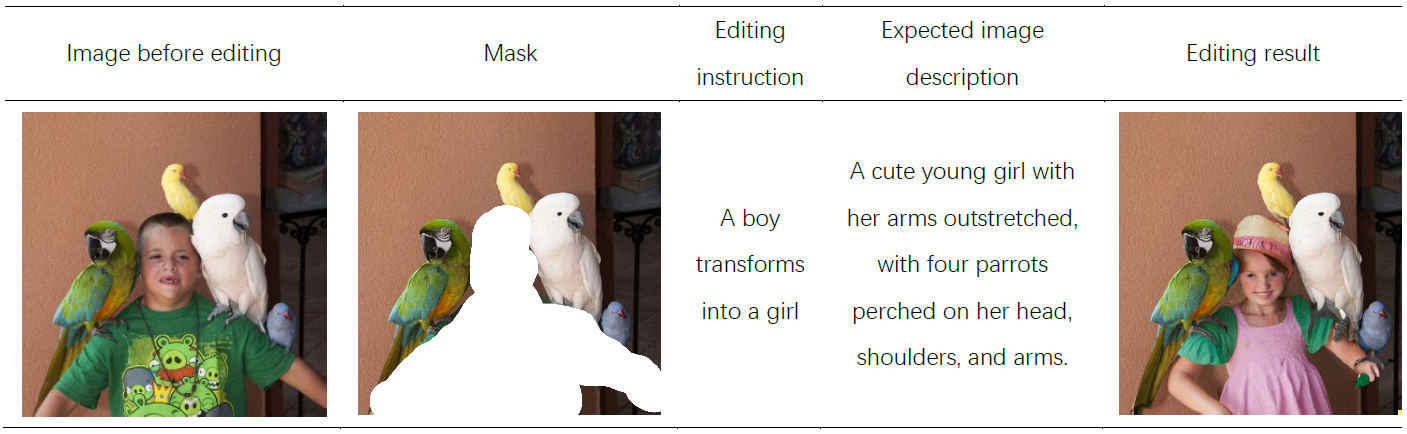}
	\vspace{-3mm}
            \caption{Submission Example.}
	\label{fig:pipe11}  
\end{figure}
\clearpage 

\section{Failure Cases (not included in \name)} \label{app:failure}
It is important to emphasize that our images underwent rigorous review and filtering. As mentioned in Section~\ref{sec:pipe}, the experts annotated approximately 20,000 images, but only 5,751 images were retained in the final \name. In this section, we present some common failure cases encountered during our data validation process. Additional examples can be found in Appendix~\ref{app:book}.

\subsection{Inherent Limitations of DALL-E 2}
The image generation rate of DALL-E 2 is relatively low, and we have identified several inherent limitations. 

\begin{figure}[h!]
	\centering  
	\includegraphics[width=1.0\textwidth]{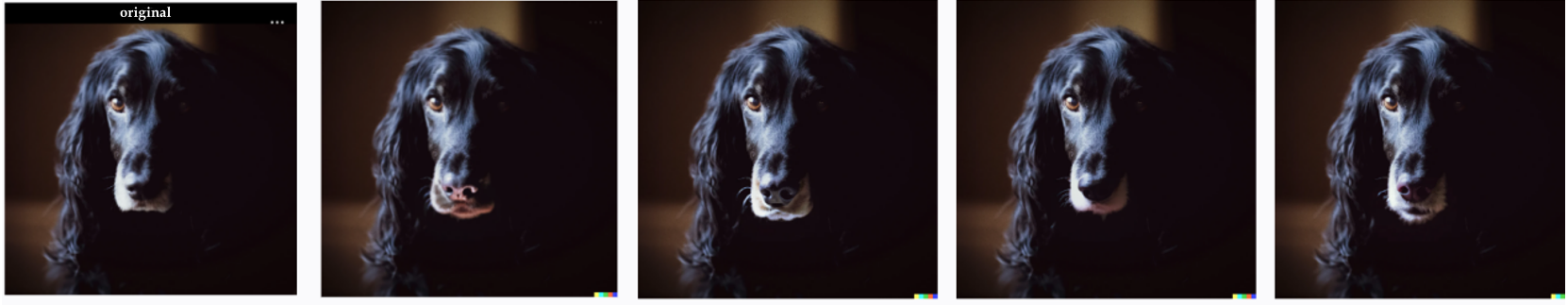}
	\vspace{-4mm}
	\caption{An Illustration of the Mismatch Between Editing Results and Instructions.}
	\label{fig:fail1}  
\end{figure}

\begin{figure}[h!]
	\centering  
	\includegraphics[width=1.0\textwidth]{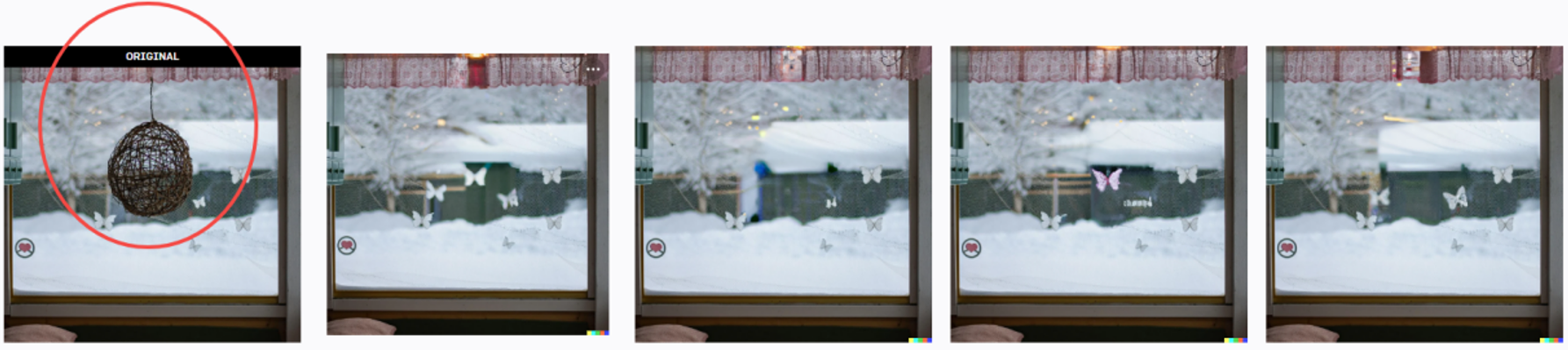}
	\vspace{-4mm}
	\caption{An Illustration of the Mismatch Between Editing Results and Instructions.}
	\label{fig:fail2}  
\end{figure}

\begin{figure}[h!]
	\centering  
	\includegraphics[width=1.0\textwidth]{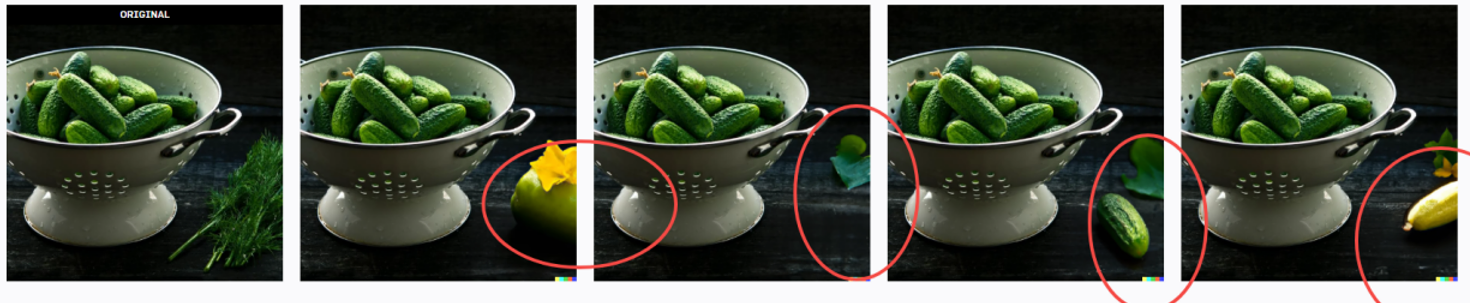}
	\vspace{-4mm}
	\caption{An Illustration of the Mismatch Between Editing Results and Instructions.}
	\label{fig:fail3}  
\end{figure}

\textbf{Mismatch between editing results and instructions.} 
For example, in Figure~\ref{fig:fail1}, the instruction was \textit{"make the nose larger,"} but no modification was applied. In Figure~\ref{fig:fail2}, the instruction was \textit{"a lantern hanging in front of the window,"} but DALL-E 2 simply removed the original object without replacing it. In Figure~\ref{fig:fail3}, the instruction was \textit{"a plate of cucumbers and a bouquet of roses,"} but the roses did not appear.


\begin{figure}[h!]
	\centering  
	\includegraphics[width=1.0\textwidth]{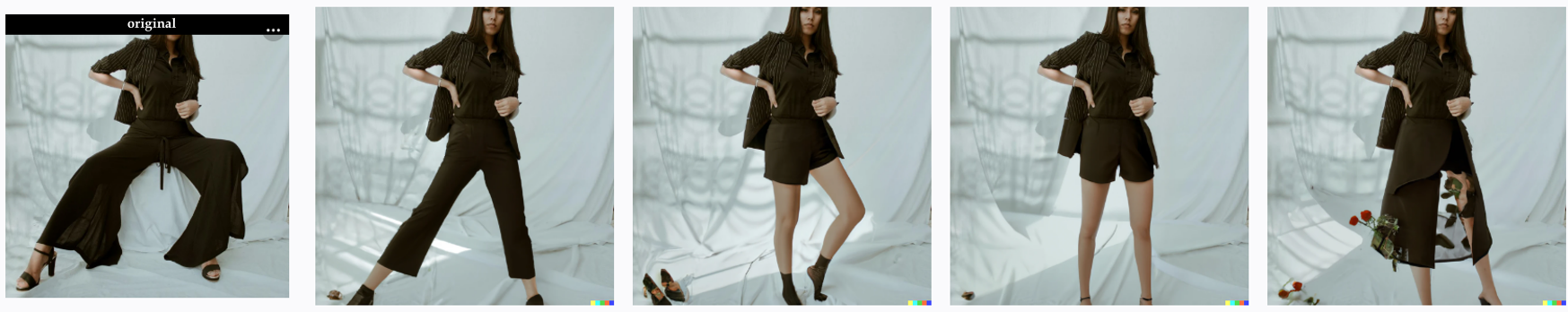}
	\vspace{-4mm}
	\caption{An Illustration of the Limited Editing Capabilities for Specific Types.}
	\label{fig:fail4}  
\end{figure}

\begin{figure}[h!]
	\centering  
	\includegraphics[width=1.0\textwidth]{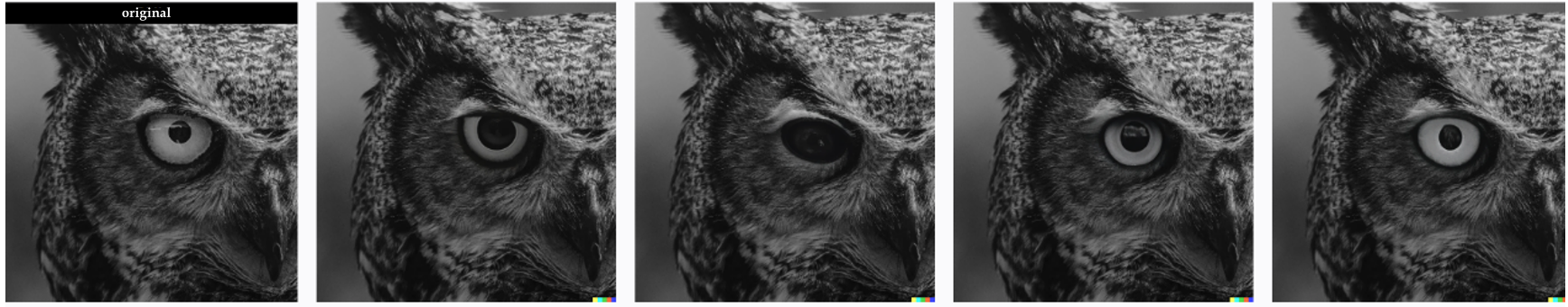}
	\vspace{-4mm}
	\caption{An Illustration of the Limited Editing Capabilities for Specific Types.}
	\label{fig:fail11}  
\end{figure}

\textbf{Limited Editing Capabilities for Specific Types}. DALL-E 2 exhibits limited performance in editing certain types of content, such as counting and relational editing tasks. Similar limitations are observed in other models as well. For instance, in Figure~\ref{fig:fail4}, the editing instruction "the girl is standing on tiptoe" was attempted multiple times by the experimental team, but a satisfactory result could not be achieved despite dozens of trials. A similar issue is seen in Figure~\ref{fig:fail11}, where the editor intended to close the owl's eyes, but DALL-E 2 continuously altered the state of the owl's eyes without successfully achieving the desired effect.

\begin{figure}[h!]
	\centering  
	\includegraphics[width=1.0\textwidth]{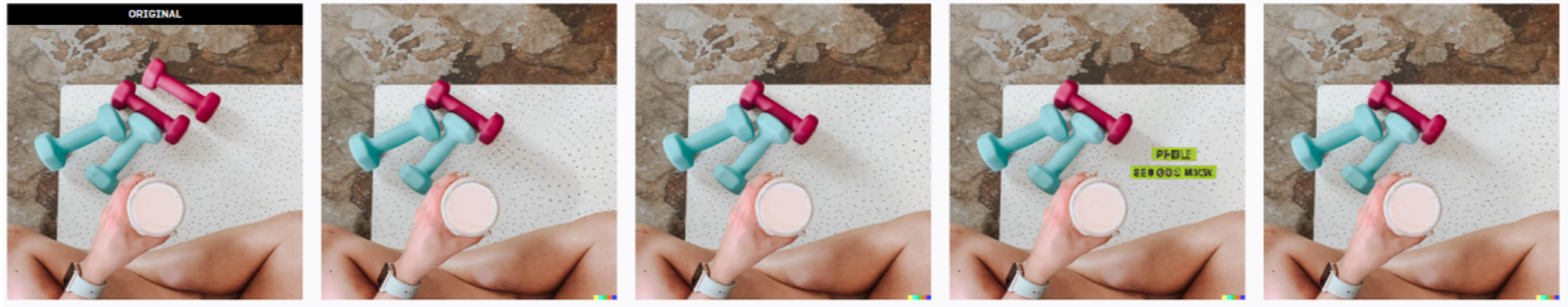}
	\vspace{-4mm}
	\caption{An Illustration of the Limited Editing Capabilities for Specific Types.}
	\label{fig:fail5}  
\end{figure}

\begin{figure}[h!]
	\centering  
	\includegraphics[width=1.0\textwidth]{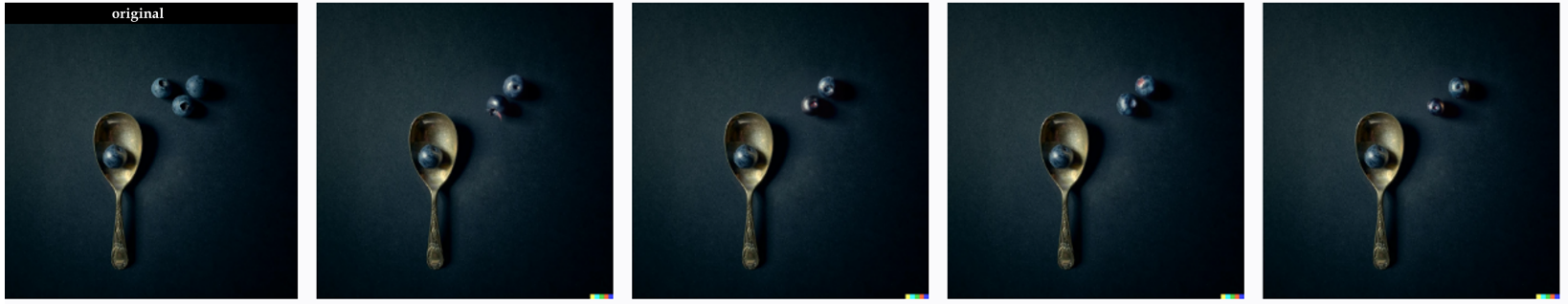}
	\vspace{-4mm}
	\caption{An Illustration of the Limited Editing Capabilities for Specific Types.}
	\label{fig:fail10}  
\end{figure}

\clearpage 
\vspace{8mm}
In the example shown in Figure~\ref{fig:fail5}, the goal was to \textit{"add a red barbell,"} but DALL-E 2 appears to be insensitive to the number of objects, with the resulting images mostly~\citep{podell2023sdxl, ge2024demon24} showing a reduction in the number of objects rather than an addition. In Figure~\ref{fig:fail10}, the editor intended to move the blueberry from the top right corner of the spoon to the top left corner, but this attempt also failed. The issue of removing rather than adding objects seems to be a common challenge across most models and may represent a significant current limitation.


Below are additional examples where DALL-E 2 fails to perform effective editing. In such cases, annotators may need to try multiple attempts and adjust the masked regions, as generation is limited to those areas. The instruction for Figure 1 is \textit{"A young boy wearing a beret"}; the instruction for Figure 2 is \textit{"A girl sitting far from the computer, pointing at it"}; and the instruction for Figure 3 is \textit{"A man raising his left fist."}
\clearpage 
\begin{figure}[h!]
	\centering  
	\includegraphics[width=1.0\textwidth]{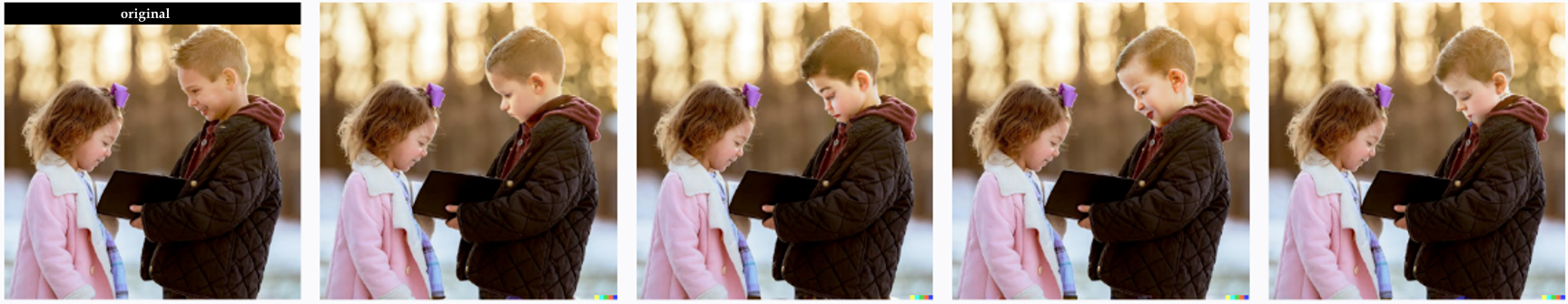}
	\vspace{-3mm}
	\caption{An Illustration of the Limited Editing Capabilities for Specific Types.}
	\label{fig:fail12}  
\end{figure}
\begin{figure}[h!]
	\centering  
	\includegraphics[width=1.0\textwidth]{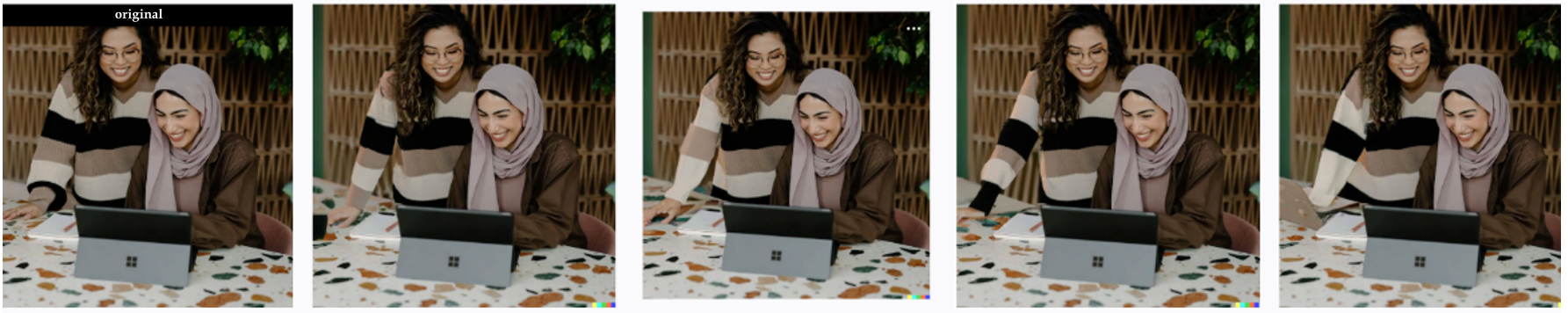}
	\vspace{-3mm}
	\caption{An Illustration of the Limited Editing Capabilities for Specific Types.}
	\label{fig:fail13}  
\end{figure}
\begin{figure}[h!]
	\centering  
	\includegraphics[width=1.0\textwidth]{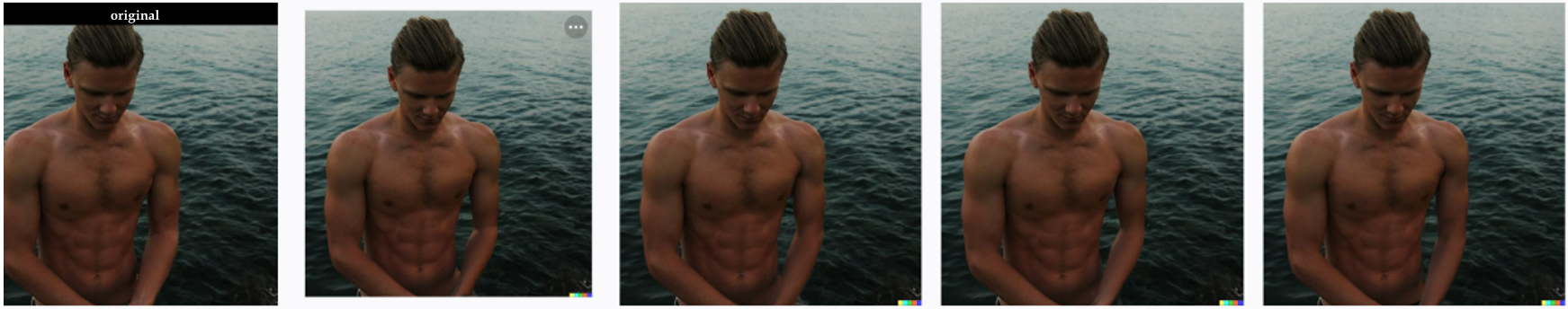}
	\vspace{-3mm}
	\caption{An Illustration of the Limited Editing Capabilities for Specific Types.}
	\label{fig:fail14}  
\end{figure}

\subsection{Editing Errors}
\begin{figure}[h!]
	\centering  
	\includegraphics[width=.82\textwidth]{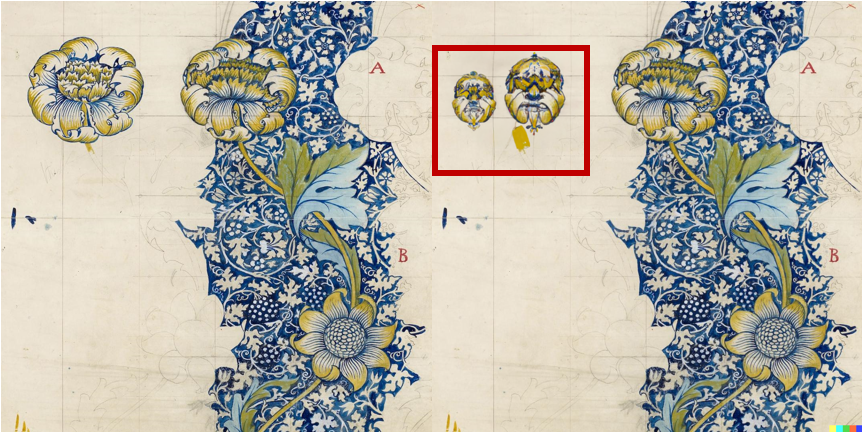}
	\vspace{-2mm}
	\caption{An example of object distortion.}
	\label{fig:fail6}  
\end{figure}

\begin{figure}[h!]
	\centering  
	\includegraphics[width=.82\textwidth]{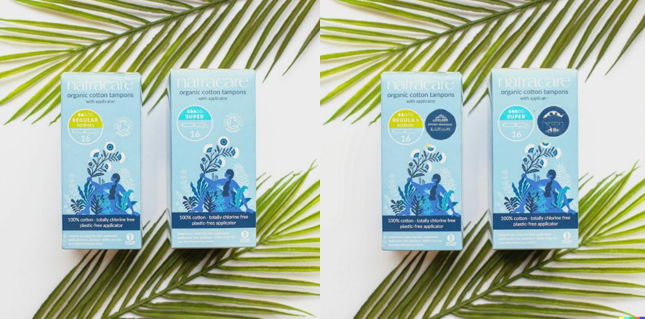}
	\vspace{-2mm}
	\caption{The discrepancy between the instruction and the generated image.}
	\label{fig:fail7}  
\end{figure}

\begin{figure}[h!]
	\centering  
	\includegraphics[width=.82\textwidth]{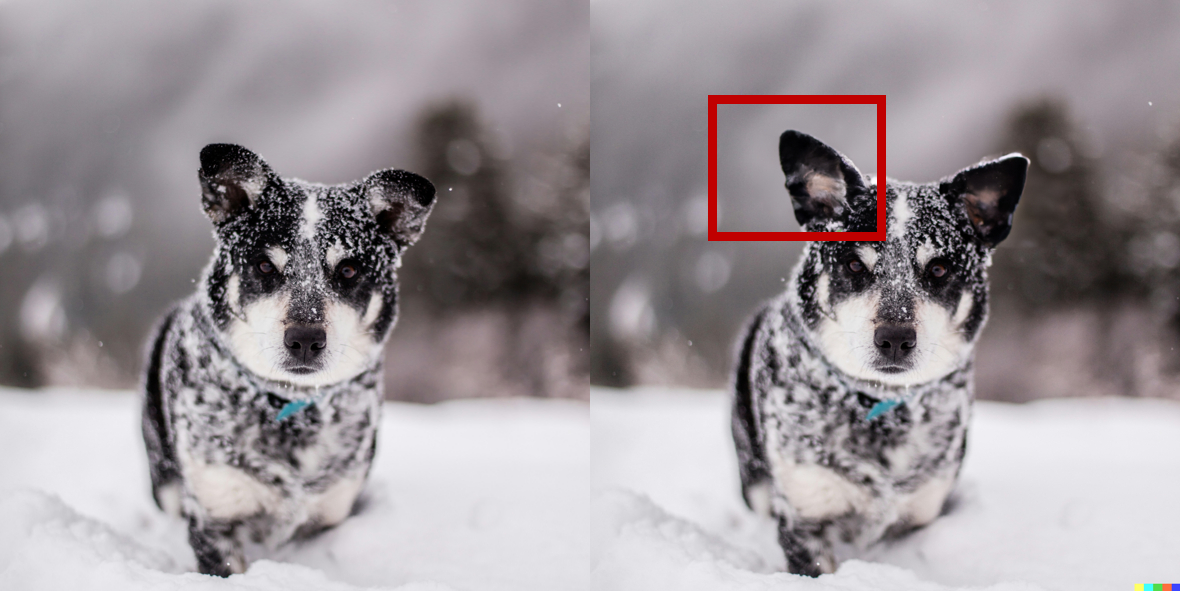}
	\vspace{-2mm}
	\caption{An example of subtle editing effects.}
	\label{fig:fail8}  
\end{figure}

\begin{figure}[h!]
	\centering  
	\includegraphics[width=.82\textwidth]{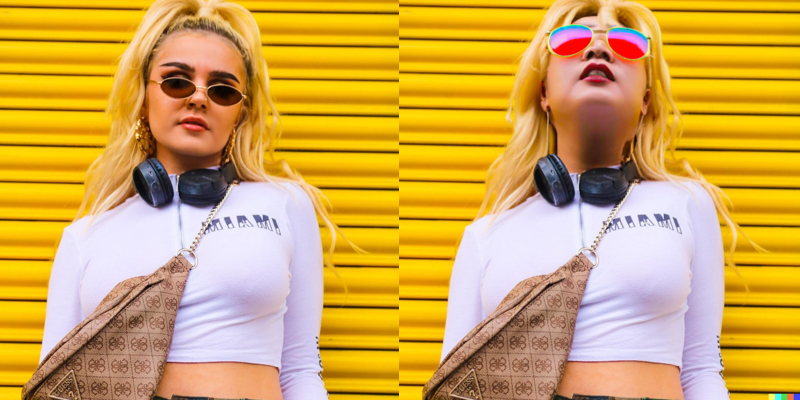}
	\vspace{-2mm}
	\caption{An example of inconsistent editing.}
	\label{fig:fail9}  
\end{figure}

Some of the editing results exhibit defects, which we have excluded from our analysis. For example, in Figure~\ref{fig:fail6}, the flower appears somewhat distorted. In Figure~\ref{fig:fail7}, the instruction is "add printed patterns," but the generated image lacks any printed patterns. In Figure~\ref{fig:fail8}, the instruction is "The puppy’s ears stood up," yet the editing effect is not clearly visible. In Figure~\ref{fig:fail9}, the instruction is to raise the person's head, but instead, the person's eyes have been altered.

\clearpage 
\subsection{Other Errors}
Additionally, we carefully reviewed the sentences in our dataset to ensure that all instructions and captions are grammatically correct and accurate. We employed a large language model~\citep{dubey2024llama} to assist in the review process, followed by manual verification. Common errors identified include:

1. The need to add "The," "A," or other determiners before nouns, such as changing "Dog raises paw" to "The dog raises its paw."
2. Incorrect pronoun references, as seen in "Move the football to the top of your feet," where "your" should be replaced with "the man's" or another appropriate description.
3. Other minor errors, such as "Lilacs change from two to one," which should be "changed" or "changes."

\clearpage

\bibliographystyle{plainnat}
\bibliography{reference}

\end{document}